\documentclass{article}
\usepackage{ijcai19}
\usepackage{times}
\usepackage{soul}
\usepackage{url}
\usepackage[hidelinks]{hyperref}
\usepackage[utf8]{inputenc}
\usepackage[small]{caption}
\usepackage{subcaption}
\usepackage{graphicx}
\usepackage{stmaryrd}
 \usepackage{mathrsfs}
 \usepackage{amssymb}
\usepackage{amsthm}
\usepackage{amsmath}
\usepackage{booktabs}
\usepackage{algorithm}
\usepackage{algorithmic}
\DeclareMathOperator*{\argmin}{arg\,min}

\usepackage{fancyhdr}
\title{A Survey of Unsupervised Domain Adaptation for Visual Recognition}

\author{
    Youshan Zhang
    \affiliations
      Computer Science and Engineering\\ Lehigh University, USA \\\emails
    yoz217@lehigh.edu
}

\begin{document}

\maketitle

\begin{abstract}
While huge volumes of unlabeled data are generated and made available in many domains, the demand for automated understanding of visual data is higher than ever before. Most existing machine learning models typically rely on massive amounts of labeled training data to achieve high performance. Unfortunately, such a requirement cannot be met in real-world applications.  The number of labels is limited and manually annotating data is expensive and time-consuming. It is often necessary to transfer knowledge from an existing labeled domain to a new domain. However, model performance degrades because of the differences between domains (\textit{domain shift} or \textit{dataset bias}). To overcome the burden of annotation, \textbf{Domain Adaptation (DA)} aims to mitigate the domain shift problem when transferring knowledge from one domain into another similar but different domain. \textbf{Unsupervised DA (UDA)} deals with a labeled source domain and an unlabeled target domain.  The principal objective of UDA is to reduce the domain discrepancy between the labeled source data and unlabeled target data and to learn domain-invariant representations across the two domains during training.  In this paper, we first define UDA problem. Secondly, we overview the state-of-the-art methods for different categories of UDA from both traditional methods and deep learning based methods. Finally, we collect frequently used benchmark datasets and report results of the state-of-the-art methods of UDA on visual recognition problem. 
\end{abstract}

\section{Introduction}
In this era of big data, huge amounts of text, images, voices, and other types of data are produced. Industry and the research community have great demand for automatic classification, segmentation, and regression for multimedia data~\cite{chen2018zero,long2014transfer}\footnote{This paper is adapted from Chapters 1 and 2 of my Ph.D. thesis: Unsupervised Domain Adaptation for
Visual Recognition~\cite{zhang2021unsupervised}.}.  Supervised learning is one of the most prevalent types of machine learning and has enjoyed much success across diverse application areas. In recent years, we have witnessed the great success of deep neural networks in some standard benchmarks such as ImageNet \cite{deng2009imagenet} and CIFAR-10 \cite{krizhevsky2010convolutional}. However, in the real world, we often have a serious problem that lacks labeled data for training. It is known that training and updating of the machine learning model depends on data annotation. Also, the high performance of machine learning models depends on the existence of massive labeled training data. Unfortunately, such a  requirement cannot be met in many real scenarios with limited or no labels of collected data. Also, a major assumption is that the training and testing data have identical distributions. Such an assumption can be easily distorted if the background, quality, or shape deformation are different across the domains.  In addition, it is often time-consuming and expensive to manually annotate data. This brings challenges to properly train and update machine learning models. As a result, some application areas have not been well developed due to insufficient labeled data for training. Therefore, it is often necessary to transfer knowledge from an existing labeled domain to a similar but different domain with limited or no labels. 

However, due to the phenomenon of data bias or domain shift~\cite{pan2010survey} (when the target distribution, from which the test images are sampled, is different from the training source distribution), machine learning models do not generalize well from an existing domain to a novel unlabeled domain. For traditional machine learning approaches, we usually assume that training data (source domain) and test data (target domain) are from the same distribution, and models are optimized from training data to directly apply in test data for prediction. The differences between training and test data are omitted. However, there are often differences between the source and target domains, and traditional approaches have lower performance if there is a domain shift issue. It is hence important to mitigate the domain shift problem to improve model performance across different domains. 

\begin{figure}[h]
\centering
\includegraphics[scale=0.6]{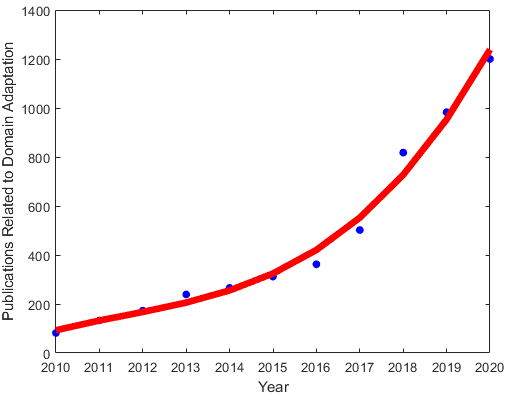}
\caption{The popularity of domain adaptation. Statistics is from searching key word ``domain adaptation" on Google Scholar (rough estimation, image from~\protect\cite{zhang2021unsupervised}).}
\label{fig:pop}
\end{figure}

Domain adaptation (DA) is one of the special settings of transfer learning (TL), which aims to leverage knowledge from an abundant labeled source domain to learn an effective predictor for the target domain with limited or no labels while mitigating the domain shift problem. In recent years, DA keeps gaining attention in the computer vision field, as shown in Fig.~\ref{fig:pop}. More and more DA related papers are published every year, which shows the importance of applications of DA. There are three types of DA (supervised, semi-supervised, and unsupervised DA), which depend on the number of labels in the target domain. For supervised DA, all target data labels are available. For semi-supervised DA, a portion of target data labels are available.  For unsupervised domain adaptation (UDA), there is no label for the target domain. To circumvent the limitations posed by insufficient annotation,  techniques combine the labeled source domain with unlabeled samples from the target domain. In addition, the number of categories of source and target domains are the same in UDA, which is also called closed set domain adaptation. 

Existing domain adaptation methods assume that the data distributions of the source and target domains are different, but share the same label space. Traditional DA methods highly depend on the extracted features from raw images.  With the development of deep neural networks, researchers are utilizing higher performance deep features (e.g., AlexNet~\cite{krizhevsky2012imagenet},  ResNet50~\cite{long2017deep}, Xception~\cite{zhang2019transductive}, InceptionResNetv2~\cite{zhang2019modified}) instead of lower-level SURF features. However, the predictive accuracy of traditional methods is affected by the quality of the extracted features from deep neural networks~\cite{zhang2020impact}. Recently, deep neural network methods witness great success in domain adaptation problems. Especially, adversarial learning shows its power in embedding in deep neural networks to learn feature representations to minimize the discrepancy between the source and target domains~\cite{tzeng2017adversarial,liu2019transferable}. However, it narrowly focuses on improving existing solutions from the source domain to the target domain, while structure information from target samples is hard to preserve. Also, it is difficult to remove noisily predicted labels in the target domain.

There have been developed several surveys on the TL and DA over the past fewer years \cite{pan2010survey,shao2015transfer,day2017survey,patel2015visual,zhang2017transfer,csurka2017domain,wang2018deep}. 
Pan and Yang \cite{pan2010survey} were the first to categorize TL under three settings: inductive TL, transductive TL, and unsupervised TL. Their focus is on the homogeneous feature spaces. Shao
et al. \cite{shao2015transfer} considered TL techniques for transferring knowledge of feature-representation level and classifier-level. Patel et al. \cite{patel2015visual} only focused on DA as a special case of TL. Day and Khoshgoftaar \cite{day2017survey} discussed heterogeneous TL in different settings. Zhang et al. \cite{zhang2017transfer} summarized different transferring criteria based on concepts of DA. In general, these five surveys only covered models on  traditional TL or DA. Later, Csurka
\cite{csurka2017domain} analyzed the state-of-the-art traditional DA methods and
categorized the deep DA method. However, Csurka’s work discussed a few deep DA methods. Wang and Deng \cite{wang2018deep} then classified the Deep DA into three groups: discrepancy based, adversarial based and reconstruction based methods based on Csurka’s work. However, they did not provide information regarding traditional methods.

In this paper, we mainly focus on the domain adaptation on image recognition tasks. The contributions of this survey are as follows. \textit{(i)} We present a taxonomy of different DA using traditional and deep learning based methods.
\textit{(ii)} We are the first who study the traditional techniques in three different settings:   feature selection, distribution adaptation, and subspace learning.
\textit{(iii)} We also discuss the deep learning based methods from discrepancy-based, adversarial-based, pseudo-labeling-based, reconstruction-based, representation-based, and attention-based methods.
\textit{(iv)} We collect several benchmark datasets, which is widely used in UDA and report results of state-of-the-art methods. 

The rest of the paper is organized as follows: In Sections~\ref{sec:notation} and~\ref{sec:DA_theory}, we introduce the notations and generalization bound of DA problem. In Section \ref{sec:tra_me}, we review the traditional methods of UDA. In Section \ref{sec:deep_me}, we describe deep DA methods for image recognition. In Section~\ref{sec:results}, we list the benchmark datasets for DA and report the accuracy of state-of-the-art methods.

\section{Notation}\label{sec:notation}
In this section, we formally define the notation in domain adaptation. A domain $\mathcal{D}$ consists of a feature space $\mathcal{X}$ by considering the marginal probability $P(\mathcal{X})$, and the task is defined by the label space $\mathcal{Y}$. The conditional distribution is $P(\mathcal{Y}|\mathcal{X})$, and the joint distribution is denoted as $P(\mathcal{X},\mathcal{Y})$.

When considering unsupervised domain adaptation in classification, there is a source domain $\mathcal{D_S} = \{\mathcal{X}_\mathcal{S}^i, \mathcal{Y}_\mathcal{S}^i \}_{i=1}^{\mathcal{N}_\mathcal{S}}$ of $\mathcal{N}_\mathcal{S}$ labeled samples in $C$ categories and a target domain $\mathcal{D_T} = \{\mathcal{X}_\mathcal{T}^j\}_{j=1}^{\mathcal{N}_\mathcal{T}}$ of $\mathcal{N}_\mathcal{T}$ samples without any labels ($\mathcal{Y}_\mathcal{T}$ is unknown), also in $C$ categories. The samples $\mathcal{X_S}$ and $\mathcal{X_T}$  obey the marginal distribution of $P(\mathcal{X_S})$ and $P({\mathcal{X_T}})$. The conditional distributions of the two domains are denoted as $P(\mathcal{Y_S}|\mathcal{X_S})$ and $P(\mathcal{Y_T}|\mathcal{X_T})$, respectively. Due to the difference of the two domains, the distributions are assumed to be different, \textit{i.e.,}  $P(\mathcal{X_S}) \neq P({\mathcal{X_T}})$ and $P(\mathcal{Y_S}|\mathcal{X_S}) \neq P(\mathcal{Y_T}|\mathcal{X_T})$. The goal for UDA is to learn  a classifier with lower generalization error in the target domain by mitigating the domain discrepancy.

\begin{figure*}[t]
\centering
\includegraphics[scale=0.3]{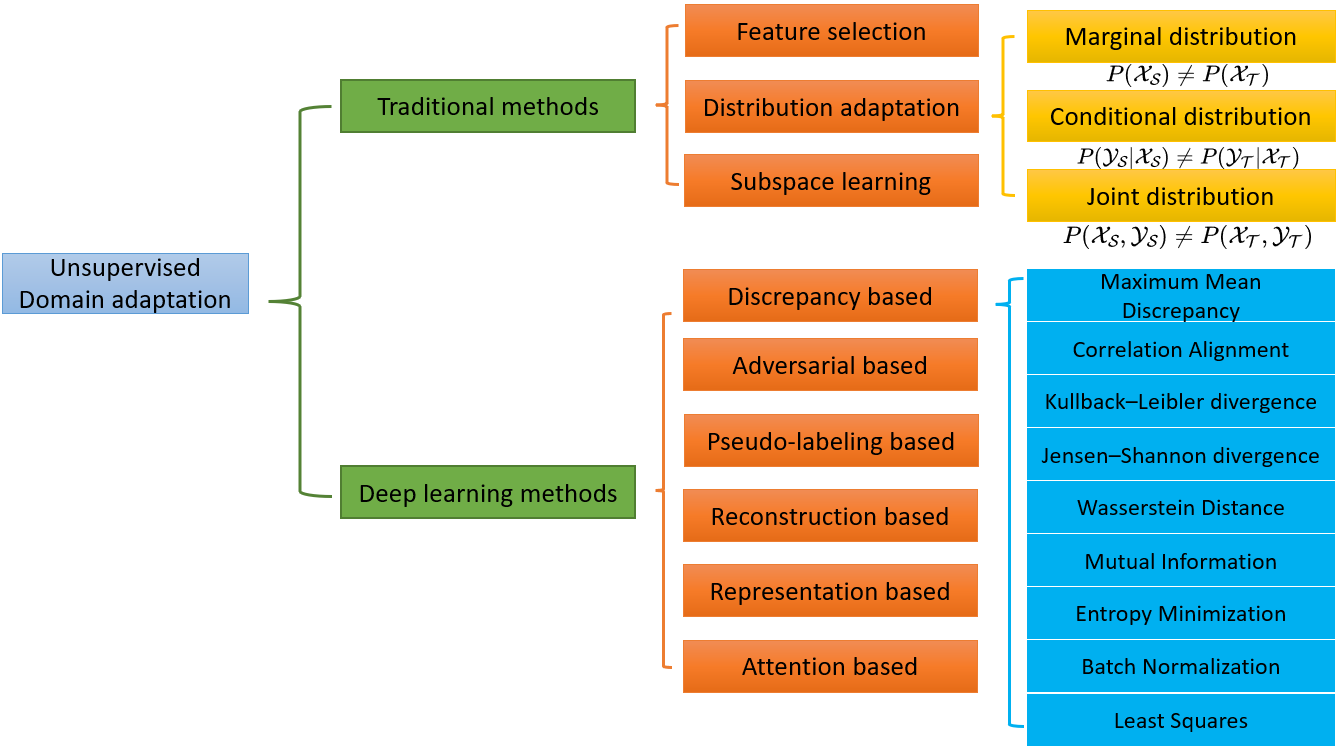}
\caption{Taxonomy of unsupervised domain adaptation for image classification task (image adapted from~\protect\cite{zhang2021unsupervised}).}
\label{fig:tox}
\end{figure*}

\section{Generalization Bound for Domain Adaptation}\label{sec:DA_theory}
Before discussing the domain adaptation methods, we first show the learning theory from Ben-David et al.~\cite{ben2010theory} to estimate the error bound of DA. It indicates that the target domain error can be minimized via bounding the source domain error and the discrepancy between them.  Differing from most conventional machine learning methods, the domain adaptation approaches not only optimize the model with the source domain, but also consider the target data and reduce the discrepancy between them in the following Theorem.

\textbf{Theorem 1} \textit{Let $\mathcal{H}$ be a hypothesis space.  Given two domains $\mathcal{D_S}$ and $\mathcal{D_T}$,  we have
\begin{equation*}
\begin{aligned}\label{eq:ana}
\forall h \in \mathcal{H},\  R_{\mathcal{T}} (h)  & 
\leq R_{\mathcal{S}} (h) + d_{\mathcal{H}\Delta\mathcal{H}}(\mathcal{D_S}, \mathcal{D_T}) + \beta, 
\end{aligned}
\end{equation*}
where $R_{\mathcal{S}} (h)$ and $R_{\mathcal{T}} (h)$ represent the source and target domain error, respectively. $d_{\mathcal{H}\Delta\mathcal{H}}$ is the discrepancy distance between two distributions $\mathcal{D_S}$ and $\mathcal{D_T}$ w.r.t.\ a hypothesis set $\mathcal{H}$. $\beta = \argmin_{h\in\mathcal{H}} R_{\mathcal{S}} (h^{*}, f_\mathcal{S}) + R_{\mathcal{T}} (h^{*}, f_\mathcal{T})$  where $f_\mathcal{S}$ and $f_\mathcal{T}$ are the label functions of the source and target domains, which can be determined by $\mathcal{Y_S}$ and pseudo target domain labels. $h^*$ is the ideal hypothesis and $\beta$ is the shared error and is expected to be negligibly small and can be disregarded.}

Recall that $R_{\mathcal{S}} (h)$ can be minimized via training the labeled source domain. Existing DA models always aim to find a minimal $d_{\mathcal{H}\Delta\mathcal{H}}(\mathcal{D_S}, \mathcal{D_T})$ to pursue a lower generalization bound of $R_{\mathcal{T}} (h)$.

According to the similarities and differences between feature space and label space, the DA can be classified into two categories: homogeneous DA and heterogeneous DA. In homogeneous DA, the feature space is the same ($\mathcal{F_S}=\mathcal{F_T}$) with the same feature dimensionality ($d_\mathcal{S}=d_\mathcal{T}$). In heterogeneous DA ($\mathcal{F_S} \neq \mathcal{F_T}$), the feature dimensionality is different ($d_\mathcal{S} \neq d_\mathcal{T} $). In this paper, we will mainly discuss homogeneous DA and focus on the most challenges of unsupervised DA. In Secs.~\ref{sec:tra_me} and~\ref{sec:deep_me}, we introduce a taxonomy of unsupervised domain adaptation for image classification task in two tracks: traditional methods and deep learning-based methods as shown in Fig.~\ref{fig:tox}.

\section{Traditional methods}\label{sec:tra_me}
In this section, we review traditional DA methods, which rely on extracted features from raw images.  As shown in Fig.~\ref{fig:tox}, we classify traditional DA methods into three sub-groups: feature selection, distribution adaptation, and subspace learning. For feature selection methods, we first learn a method to represent images, and we assume that the source and target domains share similarities in the features. Our goal is to select these features that are shared between the two domains. For distribution adaptation, we assume that the distributions of the source domain and target domain are different but share similarity, and we aim to align the distributions between the source domain and the target domain. For subspace learning, we assume that there is a shared subspace (a lower-dimensional representation) between two domains, and domain shift can be minimized in such a common subspace.


\begin{figure}[h]
\centering
\includegraphics[scale=0.35]{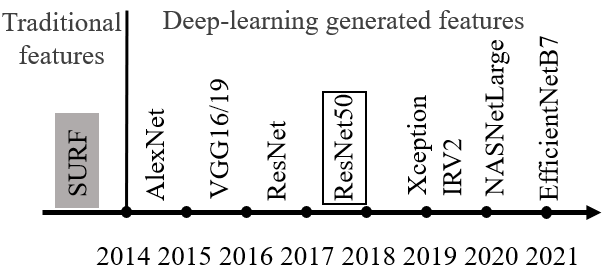}
\caption{Frequently used image feature types for DA, while ResNet50 is the most frequently used deep network for feature extraction. IRV2: InceptionResNetv2. The shading SURF is traditional feature (image from~\protect\cite{zhang2021unsupervised}).}
\label{fig:fea}
\end{figure}

\subsection{Feature selection methods}
The first step for visual recognition is to find a proper way to represent images. In recent decades, with the emergence of deep networks, the feature representation of images has changed rapidly.  As shown in Fig.~\ref{fig:fea}, speeded up robust features (SURF)  is one of the most popular extracted features for visual recognition before the deep features. It is a fast and robust algorithm for local, similarity invariant representation, and comparison of images feature~\cite{bay2006surf}. However, SURF can only detect some points, but not all important features. After the emergence of different ImageNet-trained deep models, their deep features have been widely used in the field of computer vision as shown in Fig.~\ref{fig:fea}. The underlying assumption of the feature selection method is that both the source domain and the target domain contain  at least some common features. The goal of this kind of method is to select these shared features through a machine learning method and then build models based on these features.

Structural correspondence learning (SCL)~\cite{blitzer2006domain} is one of the most representative models to find the common features of both domains. These common features are named as Pivot features, which refer to the words that frequently appear in different domains in text classification. Due to the stability of these features, they can be used as the bridge to transfer knowledge. It has three steps. 1) Feature Selection: SCL first obtains the pivot features;
2) Mapping Learning: the pivot features are utilized to
find a low-dimensional common latent feature space;
3) Feature Stacking: a new feature representation is constructed by feature augmentation.

\begin{figure}[h]
\centering
\includegraphics[scale=0.23]{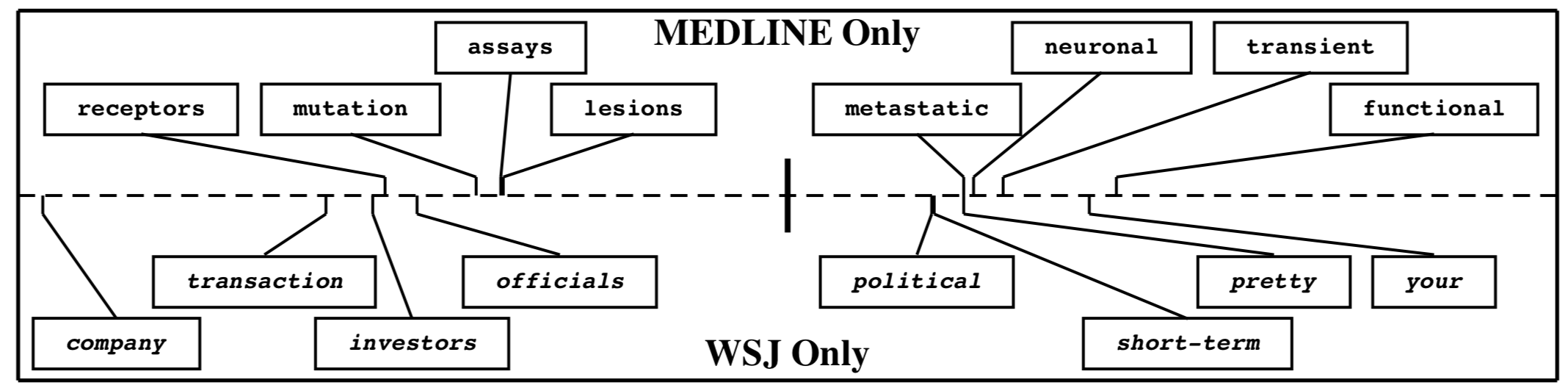}
\caption{The scheme of Pivot feature in feature selection methods (image from \protect\cite{blitzer2006domain}). }
\label{fig:SCL}
\vspace{-0.3cm}
\end{figure}

Esmat et al.~\cite{rashedi2013simultaneous} proposed a mixed gravitational search algorithm (MGSA) to reduce the semantic gap between low-level visual features and high-level semantics through simultaneous feature adaptation and feature selection. Later, feature selection and structure preservation (FFSL) \cite{li2016joint} smoothly integrated structure preservation and feature selection into a unified optimization problem. They first selected relevant features across two domains and then utilized a nearest neighbor graph and a representation matrix to preserve the geometric structure. Also, there are extended works to incorporate other techniques. Gu et al.~\cite{gu2011joint} proposed a joint feature selection and a subspace learning model to unify feature selection and subspace learning in a  framework. Transfer Joint Matching (TJM)~\cite{long2014transfer}, simultaneously adapted marginal distribution and performed source domain sampling selection during the process of optimizing an objective function. Combining deep features with traditional methods has also been explored~\cite{zhang2018automated,zhang2019regressive,wu2020mixing}, Zhang et al. investigated how different pre-trained ImageNet models affect transfer accuracy on domain adaptation problems~\cite{zhang2020impact}. They found that features from a better ImageNet model can improve the performance of domain adaptation. This observation was further validated by their later work~\cite{zhang2021efficient}.

\begin{figure}[h]
\centering
\includegraphics[scale=0.29]{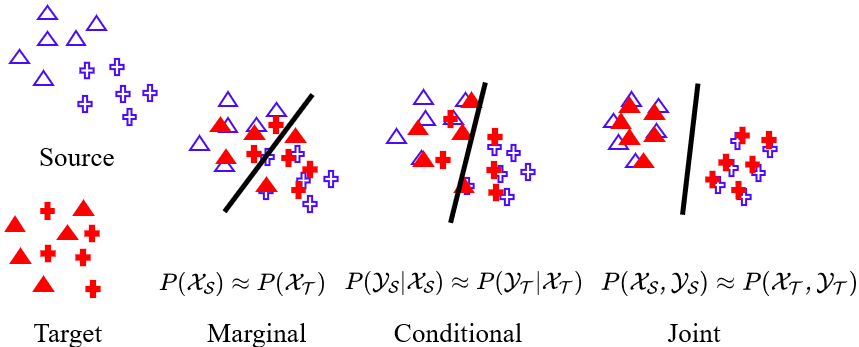}
\caption{An example of different types of distribution alignment. Type I: marginal distribution; Type II: conditional distribution; Type III: joint distribution (including aligning both I and II). Black line: classifier (image from~\protect\cite{zhang2021unsupervised}). }
\label{fig:CATE}
\end{figure}

\subsection{Distribution adaptation methods}
Distribution adaptation methods can be classified into three categories: marginal distribution adaptation ($P(\mathcal{X_S})\neq$  $P(\mathcal{X_T})$), conditional distribution adaptation ($P(\mathcal{Y_S}|\mathcal{X_S})\neq$  $P(\mathcal{Y_T}|\mathcal{X_T})$) and joint distribution adaptation ($P(\mathcal{X_S},\mathcal{Y_S})\neq 
 P(\mathcal{X_T}, \mathcal{Y_T})$). Therefore, many methods aim to minimize domain shift from these three directions to make these distributions are similar to each other across different domains.  Fig.~\ref{fig:CATE} illustrates the different priorities of distribution, Type I first aligns the marginal distribution and type II first aligns the conditional distribution, and type III aligns the marginal and conditional distribution together.

\subsubsection{Marginal distribution adaptation} In this setting, it assumes that the marginal distributions between the two domains are different ($P(\mathcal{X_S)}\neq$  $P\mathcal{(X_T})$), which should be aligned first. This pattern is shown as the marginal distribution alignment in Fig.~\ref{fig:CATE}, which focused on overall shape alignment. It minimizes the distance between the probabilistic distribution of source and target domain in Eq.~(\ref{eq:marginal}).
\begin{equation}\label{eq:marginal}
 \min d_{\mathcal{H}\Delta\mathcal{H}}(\mathcal{D_S}, \mathcal{D_T}) \approx ||P(\mathcal{X_S})-P(\mathcal{X_T})||,
\end{equation}
where $d_{\mathcal{H}\Delta\mathcal{H}}$ is the discrepancy distance between the two domains ($\mathcal{D_S}$ and $\mathcal{D_T}$), $||\cdot||$ is the L2 norm.

\begin{figure}[h]
\centering
\includegraphics[scale=0.42]{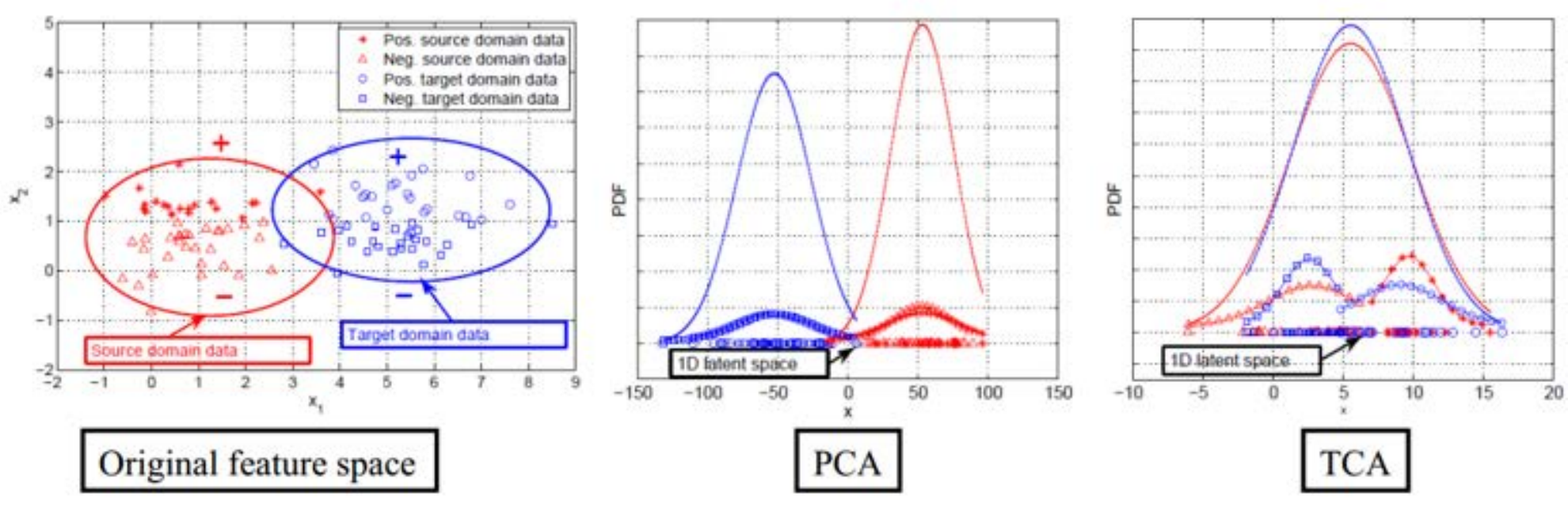}
\caption{Comparison of TCA and PCA. The red color is the source distribution and blue color is the target distribution. The distribution between two domains are more closed to each other after preforming TCA (image modified from \protect\cite{pan2011domain}). }
\label{fig:TCA}
\end{figure}

Maximum mean discrepancy (MMD) is one of most classical measurements to align the data distribution of the two domains~\cite{pan2011domain,dorri2012adapting,long2014transfer}, and its distance function is defined in Eq.~(\ref{eq:mmd}). 
\begin{equation}\label{eq:mmd}
MMD(\mathcal{D_S}, \mathcal{D_T})=||\frac{1}{\mathcal{N_S}}\sum_{i=1}^{\mathcal{N_S}}\phi{(\mathcal{X}_\mathcal{S}^i})- \frac{1}{\mathcal{N_T}}\sum_{j=1}^{\mathcal{N_T}}\phi{(\mathcal{X}_\mathcal{T}^j)}||_\mathcal{H}^2,
\end{equation}
where $\mathcal{N_S}$ and $\mathcal{N_T}$ are number of samples in the source and target domain, $\phi$ is the mapping and $\mathcal{H}$ is the universal Reproducing Kernel Hilbert Space (RKHS).

Pan et al.~\cite{pan2011domain} introduced transfer component analysis (TCA) to adopt MMD and measured the marginal distribution difference in a RKHS by enforcing the scatter matrix (a statistic that can make estimates of the covariance matrix) as a constraint. TCA assumed that there is a map ($\phi$), which can make $P(\phi(\mathcal{X_S}))\approx P(\phi(\mathcal{X_T}))$. The conditional  distribution is also similar ($P(\mathcal{Y_S}|\phi(\mathcal{X_S}))\approx P(\mathcal{Y_T}|\phi(\mathcal{X_T}))$). In TCA, it learns a linear mapping from an empirical kernel feature space to a low-dimensional feature space. In this way, it has a relatively low computational burden.  

Later, there were also more proposed models based on TCA (\cite{dorri2012adapting,long2014transfer,baktashmotlagh2016distribution,jiang2017integration}). 
Adapting component analysis (ACA)~\cite{dorri2012adapting} addressed the difference of the marginal distribution by a Hilbert Schmidt independence criteria (HSIC) based on TCA.  Duan et al.~\cite{duan2012domain} proposed a unified framework termed domain transfer multiple kernel learning (DTMKL). DTMKL introduced the multiple kernel-based TCA, and the kernel function
is assumed to be a linear combination of a group of base
kernels. Then the marginal distribution can be minimized. Transfer joint matching (TJM)~\cite{long2014transfer} updated the marginal distribution while optimizing objective functions. Distribution matching embedding (DME)~\cite{baktashmotlagh2016distribution} first calculated the transformation matrix and then performed the feature map. Another method called ITCA \cite{jiang2017integration} updated the global and local marginal distributions at the same time. 

\begin{figure}[h]
\centering
\includegraphics[scale=0.23]{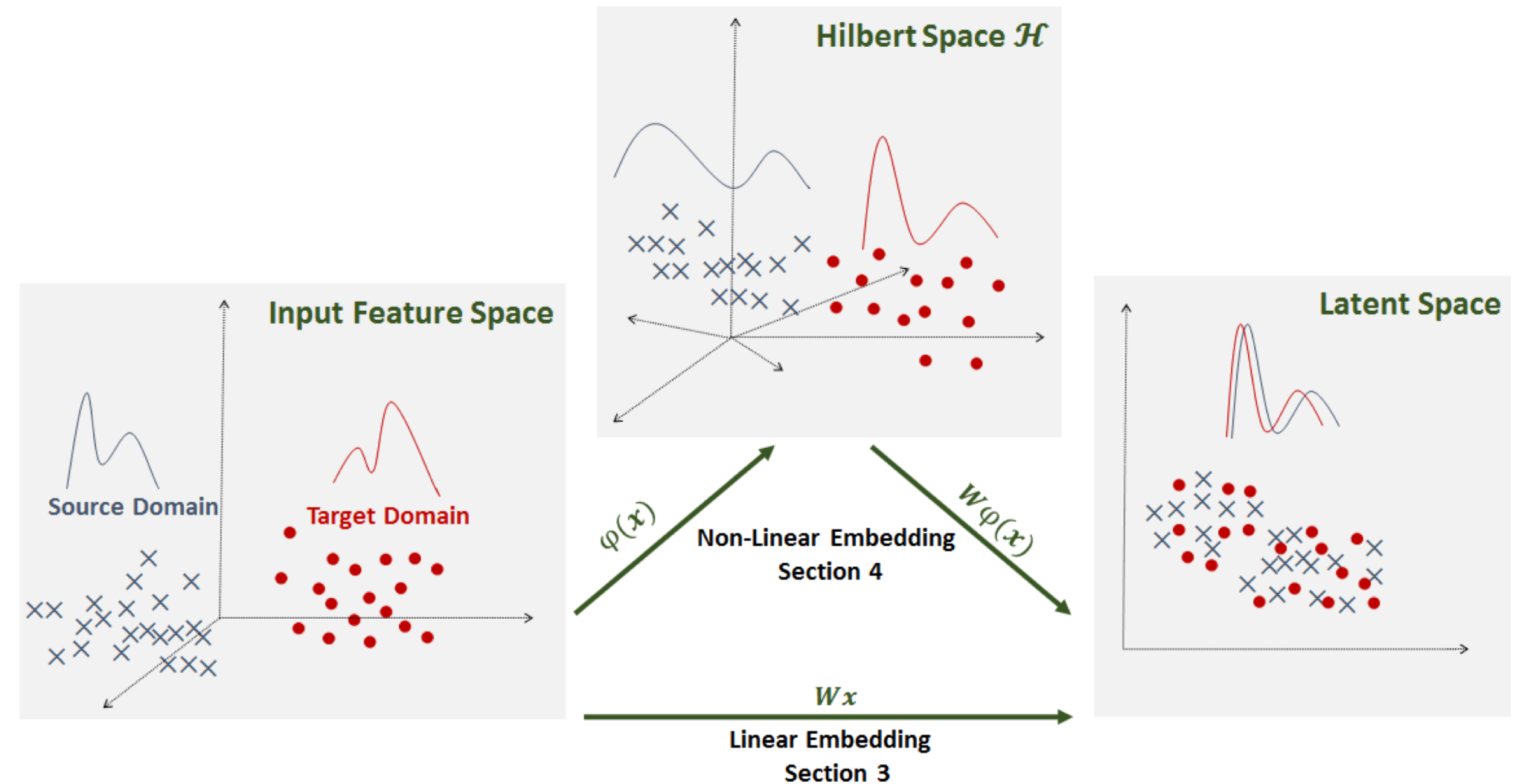}
\caption{The scheme of distribution matching embedding (DME)  model (image from \protect\cite{baktashmotlagh2016distribution}). }
\label{fig:DME}
\end{figure}

However, marginal distribution alignment assumes that the conditional distribution between two domain are similar to each other once the marginal distribution is aligned. In the real case, such an assumption is usually not valid; therefore, the conditional distribution should also be aligned. 

\subsubsection{Conditional Distribution Adaptation} 
In this setting,  we assume that the conditional distribution is varied between two domains ($P(\mathcal{Y_S}|\mathcal{X_S}) \neq  P(\mathcal{Y_T}|\mathcal{X_T})$). Many methods minimize the conditional distribution distance between the source and target domain as follows.

\begin{equation}\label{eq:conditional}
\min d_{\mathcal{H}\Delta\mathcal{H}}(\mathcal{D_S}, \mathcal{D_T}) \approx ||P(\mathcal{Y_S}|\mathcal{X_S}) -  P(\mathcal{Y_T}|\mathcal{X_T})||,
\end{equation}

Fig.~\ref{fig:CATE} also shows the conditional distribution adaptation, and it focuses on aligning the categorical distributions. However, due to the unlabeled target domain, such an alignment is difficult. Therefore, many methods take advantage of pseudo labels of the target domain.


\begin{figure}[h]
\centering
\includegraphics[scale=0.8]{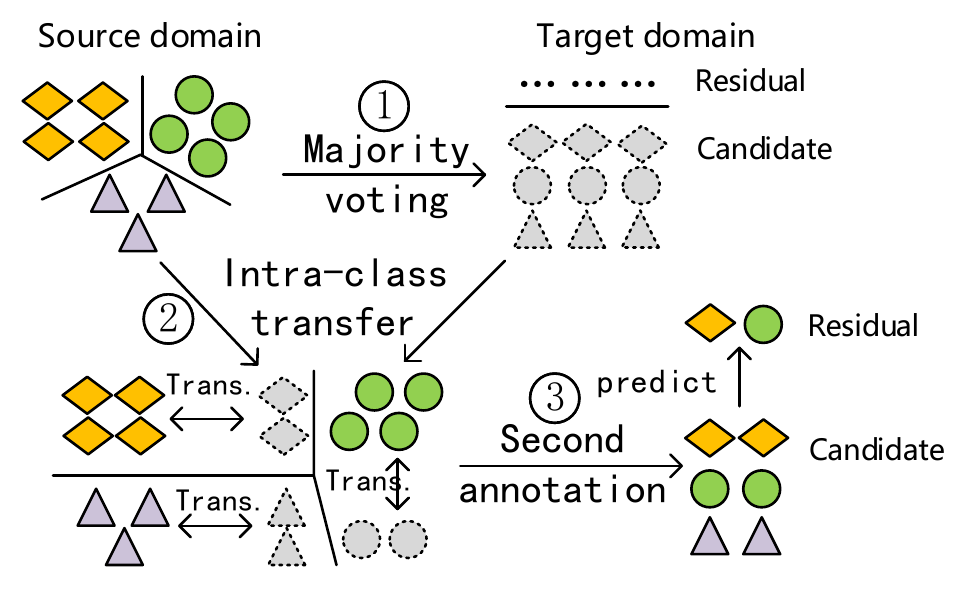}
\caption{The scheme of stratied transfer learning (STL) model
  (image from \protect\cite{gong2016domain}). }
\label{fig:STL}
\end{figure}

In this scope, Satpal and Sarawagi~\cite{satpal2007domain} proposed conditional probability models via feature subsetting. They combined conditional random fields and conditional probability adaptation to reduce the prediction error. Elsewhere~\cite{gong2016domain}, they proposed to extract conditional transferable components (CTC) from conditional distribution first, and then the marginal distribution is modeled.  Later, Wang et al. \cite{wang2018stratified} introduced the stratified transfer learning (STL) model. Most previous models are based on the global domain shift (inter-class transfer). However, it ignored intra-class transfer. Since the intra-class transfer can utilize the intra-class features, it is able to achieve a better transfer performance. The basic idea of the STL method has three steps. At the first step, majority voting is used to generate pseudo-labels for uncalibrated location behavior; then, as the next step in the reproducing kernel Hilbert space, the intra-class correlation is used to reduce the dimensionality adaptively. Note that dimensionality reduction adaptively makes the correlation between behavior data in different situations. Finally, the accurate calibration of unknown data is realized by secondary calibration. To determine the intra-class transfer, they calculated the MMD for each class using the equation given below:
\begin{equation}\label{eq:mmd_each}
Dist(\mathcal{D}_S, \mathcal{D}_T)=\sum_{c=1}^C||\frac{1}{n_{c}}\sum_{x_i \in \mathcal{D_S^C}}\phi{(x_i)}- \frac{1}{m_C}\sum_{x_j \in \mathcal{D_T^C}}\phi{(x_j)}||_\mathcal{H}^2,
\end{equation}
where $C$ represents label categories, 
$\mathcal{D_S^C}$ and $\mathcal{D_T^C}$ is the $C$ category of the source and the target domains, respectively. The STL method carried out cross-location behavior recognition experiments using a large number of behavior recognition data. 

Although the conditional distribution alignment shows a higher performance than marginal distribution alignment, it still lacks the consideration of overall shape adaptation. Therefore, joint alignment of these two distributions is necessary.

\subsubsection{Joint Distribution Adaptation} 

In this setting, many methods minimize the joint distribution distance between the source domain and the target domain in Eq.~(\ref{eq:joint}).
\begin{equation}\label{eq:joint}
\begin{aligned}
  \min d_{\mathcal{H}\Delta\mathcal{H}}(\mathcal{D_S}, \mathcal{D_T}) & \approx \ ||P(\mathcal{X_S})-P(\mathcal{X_T})|| + \\ &||P(\mathcal{Y_S}|\mathcal{X_S}) -  P(\mathcal{Y_T}|\mathcal{X_T})||,   
\end{aligned}
\end{equation}
The joint distribution adaptation corresponds to the last distribution in Fig.~\ref{fig:CATE}. Joint distribution adaptation (JDA)~\cite{long2013transfer} was proposed to reduce the marginal and conditional distributions. The main idea of the JDA method is to find a transformation $A$ to reduce the distance between $P(A^T\mathcal{X_S})$ and $P(A^T\mathcal{X_T})$. The distance between $P(\mathcal{Y_S}|A^T\mathcal{X_S})$ and $P(\mathcal{Y_T}|A^T\mathcal{X_T})$ also should be minimized. JDA model can be divided into two steps: marginal distribution adaptation and conditional distribution adaptation. For the marginal distribution adaptation, it aimed to minimize Eq.~(\ref{eq:jda_mar}).
\begin{equation}\label{eq:jda_mar}
\begin{aligned}
  ||\frac{1}{\mathcal{N_S}} \sum_{i=1}^{\mathcal{N_S}} A^T \mathcal{X}_\mathcal{S}^i -\frac{1}{\mathcal{N_T}} \sum_{j=1}^{\mathcal{N_T}}A^{T}\mathcal{X}_\mathcal{T}^j||_{\mathcal{H}}^2,   
\end{aligned}
\end{equation}

For conditional  distribution adaptation, it aimed to minimize Eq.~(\ref{eq:jda_con}).

\begin{equation}\label{eq:jda_con}
\sum_{c=1}^C||\frac{1}{n_{c}}\sum_{x_i \in \mathcal{D_S^C}}A^T x_i- \frac{1}{m_c}\sum_{x_j \in \mathcal{D_T^C}} A^T x_j||_\mathcal{H}^2,
\end{equation}

To realize it, the MMD metric and the pseudo label strategy are adopted. The desired transformation
matrix can be obtained by solving a trace optimization problem via eigen-decomposition. Further, it is obvious that the accuracy of the estimated pseudo labels affects the performance of JDA. In order to improve the labeling quality, the authors adopt iterative refinement operations. Specifically, in each iteration, JDA is performed, and then a classifier is trained on the instances with the
extracted features. Next, the pseudo labels are updated based on the trained classifier. After that, JDA is performed repeatedly with the updated pseudo labels. The iteration ends when convergence occurs.

\begin{figure}[h]
\centering
\includegraphics[scale=0.4]{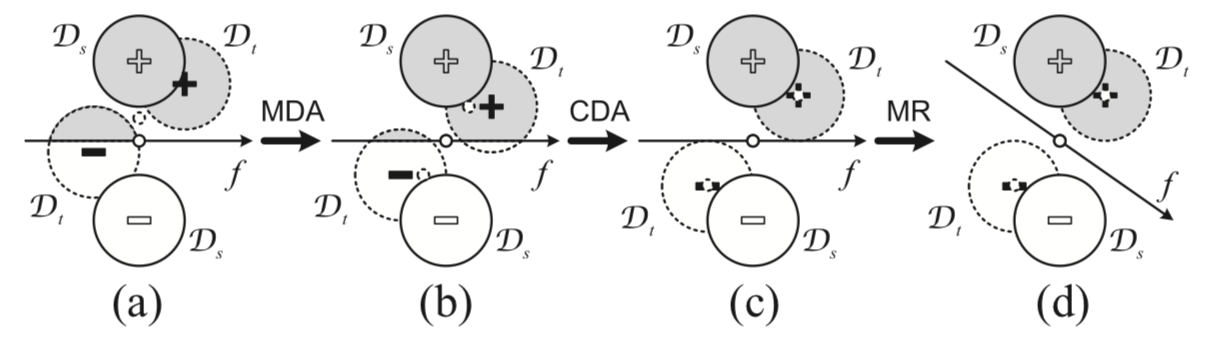}
\caption{The scheme of adaptation regularization based transfer learning (ARTL) model. MDA: marginal distribution adaptation; CDA: conditional distribution adaptation; MR: manifold regularization (image from \protect\cite{long2013adaptation}). }
\label{fig:ARTL}
\end{figure}

In follow-up work, additional loss items are added on the basis of JDA, which greatly improves the effect of transfer learning.  Adaptation regularization transfer learning (ARTL)~\cite{long2013adaptation} embedded the JDA model into a minimum structure risk framework, which represents the directed learning classifier.  The authors also proposed two specific algorithms under this framework based on different loss functions. In these two algorithms, the coefficient matrix for computing MMD and
the graph Laplacian matrix for manifold regularization are
constructed at first. Then, a kernel function is selected to
construct the kernel matrix. Fig.~\ref{fig:ARTL} illustrates the scheme of ARTL model. 


Visual domain adaptation (VDA)~\cite{tahmoresnezhad2017visual} added the intra-class and inter-class distances in the objective function based on JDA. Hsiao et al., \cite{hsiao2016learning} controlled the structure invariant based on JDA. Hou et al., proposed a model to select the target domain \cite{hou2015unsupervised}, and joint geometrical and statistical alignment (JGSA) \cite{zhang2017joint} calculated intra-class, inter-class distance, and label persistence based JDA. However, a disadvantage of JDA is that marginal distribution and conditional distribution are not equally important. Therefore, balanced distribution adaptation (BDA)~\cite{wang2017balanced} was proposed to solve this problem. The classifier $f$ keeps updating with different steps. It aimed to control the balance between two distributions via the balance factor $\mu$ using Eq.~(\ref{eq:bda}).
\begin{equation}\label{eq:bda}
\begin{aligned}
  \min d_{\mathcal{H}\Delta\mathcal{H}}(\mathcal{D_S}, \mathcal{D_T}) \approx & (1-\mu) Dist(P(\mathcal{X_S}), P(\mathcal{X_T})) \\ & +\mu Dist(P(\mathcal{Y_S}|\mathcal{X_S}), P(\mathcal{Y_T}|\mathcal{X_T})),     
\end{aligned}
\end{equation}
where $\mu \in [0, 1]$ is the balance factor; $\mu \shortrightarrow 0$ means that there is a significant difference between the source domain and the target domain data and $\mu \shortrightarrow 1$ implies that the source domain and the target domain datasets have high similarity. Wang et al. noted that conditional distribution adaptation is more important. The balance factor can dynamically adjust the importance of each distribution according to the actual data distribution and achieve a good distribution adaptation effect. When $\mu =0$, BDA is the TCA model, and if $\mu = 0.5$, the BDA becomes the JDA model. In addition, they also proposed the weighted BDA (WBDA). In WBDA~\cite{wang2017balanced}, the conditional distribution difference is measured by a weighted version of MMD to solve the class imbalance problem.

\begin{figure}[t]
\centering
\includegraphics[scale=0.450]{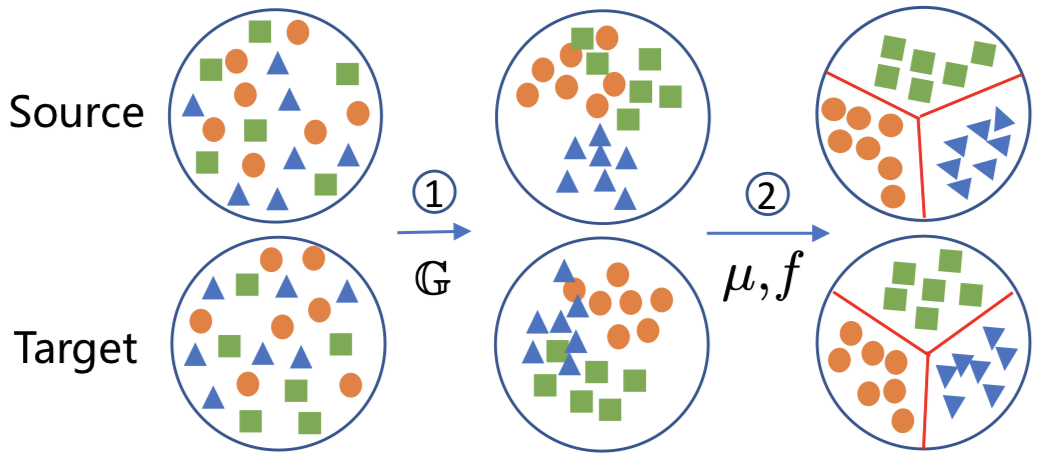}
\caption{The scheme of the MEDA model. Features are first learned via manifold kernel $G$. Then, dynamic distribution alignment will learn the domain-invariant classier $f$ (image from \protect\cite{wang2018visual}).}
\label{fig:Meda}
\end{figure}

Several approaches have addressed the alignment of marginal distribution and conditional distribution of data in special cases. Wang and Mahadevan aligned the source and target domain by preserving the ‘neighborhood structure’ of the data   points \cite{wang2009manifold}.  Wang et al.\ proposed a manifold embedding distribution alignment method (MEDA) (based on work of Gong et al.~\cite{gong2012geodesic}) to align both the degenerate feature transformation and the unevaluated distributions of both domains~\cite{wang2018visual}. The scheme of MEDA. MEDA model has three fundamental steps: 1) learn features from the manifold based on Gong et al.~\cite{gong2012geodesic}; 2) use dynamic distribution alignment to estimate the marginal and conditional distributions of data; and, 3) update the classified labels based on estimated parameters. 

The classifier ($f$) is defined as:
\begin{equation}
\begin{aligned}\label{eq:meda}
f= &\mathop{\arg\min}_{f \in  \mathcal{H}_{k}}  \sum_{i=1}^{\mathcal{N_S}} \mathop{l} (f( g(\mathcal{X}_{\mathcal{S}}^{i})),\mathcal{Y}_{\mathcal{S}}^{i})+\eta ||f||_{K}^{2} + \\ &\lambda \overline{D_{f}} (\mathcal{D_S}, \mathcal{D_T}) + \rho R_{f}(\mathcal{D_S},\mathcal{D_T} )
\end{aligned}
\end{equation}
where $\mathcal{H}_{k}$ represents kernel Hilbert space; $l(\cdot, \cdot)$ is the loss function; $g(\cdot)$ is a feature learning function in Grassmannian manifold \cite{gong2012geodesic}; $\mathcal{X_{S}}$ is the learned features from one of ImageNet models, $||f||_K^2$ is the squared norm of $f$;  $\overline{D_{f}}(\cdot, \cdot)$ represents the dynamic distribution alignment; $R_{f}(\cdot, \cdot)$ is a Laplacian regularization; $\eta$, $  \lambda$, and $\rho$ are regularization parameters. Here, the term $\mathop{\arg\min}_{f \in  \mathcal{H}_{k}}  \sum_{i=1}^{\mathcal{N_S}} \mathop{l} (f( g(\mathcal{X}_{\mathcal{S}}^{i})),\mathcal{Y}_{\mathcal{S}}^{i})+\eta ||f||_{K}^{2}$ is the source structure risk minimization (SRM). We can only employ the SRM on $\mathcal{X_S}$, since there are few labels (perhaps no labels) for $\mathcal{X_T}$.
By training the classifier from Eq.~(\ref{eq:meda}), we can predict labels of test data. Here, the balance factor $\mu$ minimizes MMD, and it can dynamically change according to the importance between source and target domain. Therefore, it achieves a higher accuracy (MEDA $>$ BDA $>$ JDA $>$ TCA $>$ conditional distribution adaptation $>$ marginal distribution adaptation~\cite{zhang2019transductive}). Zhang et al.~\cite{zhangcorrelated2021} proposed to extract both marginal and condiction features from a pre-trained ImageNet model to form the joint features and then minimize the joint distribution between the two domains based on MEDA.  

Note that most feature selection and distribution alignment methods focus on the explicit features in the original feature space.
In contrast, subspace learning also focuses on some implicit features in an underlying subspace, which can show the geometry of data. Therefore, subspace learning can play various roles in the feature
transformation process.

\subsection{Subspace learning methods}
There are two sub-categories of subspace learning models: feature alignment and manifold learning. Feature alignment methods aim to align the source feature with target features. 
One of the earliest subspace learning methods is called subspace alignment (SA) \cite{fernando2013unsupervised}. It can align the source domain and target domain via PCA with a lower subspace dimensionality $d$, which is determined by the minimum Bregman divergence of two subspaces  and it minimizes the following function:

\begin{equation}\label{eq:SA}
 F(M)=||\mathcal{X}_\mathcal{S}^{'} M-\mathcal{X}_\mathcal{T}^{'}||_{F}^{2}, 
\end{equation}
where $||\cdot||_{F}^{2}$ is the Frobenius norm and $M$ is the transformation matrix, and $\mathcal{X}_\mathcal{S}^{'} \in \mathbb{R}^{\mathcal{N_S}\times d}$ and $\mathcal{X}_\mathcal{T}^{'} \in \mathbb{R}^{\mathcal{N_T}\times d}$ are generated from the first $d$ eigenvectors from the original domain data ($\mathcal{X}_\mathcal{S}^{'}$ and $\mathcal{X}_\mathcal{T}^{'}$ are the representations of the source and target data in the reduced dimensionality subspace). Then, a learner can be trained on
the transformed matrix $F(M)$.

\begin{figure}[h]
\centering
\includegraphics[scale=0.35]{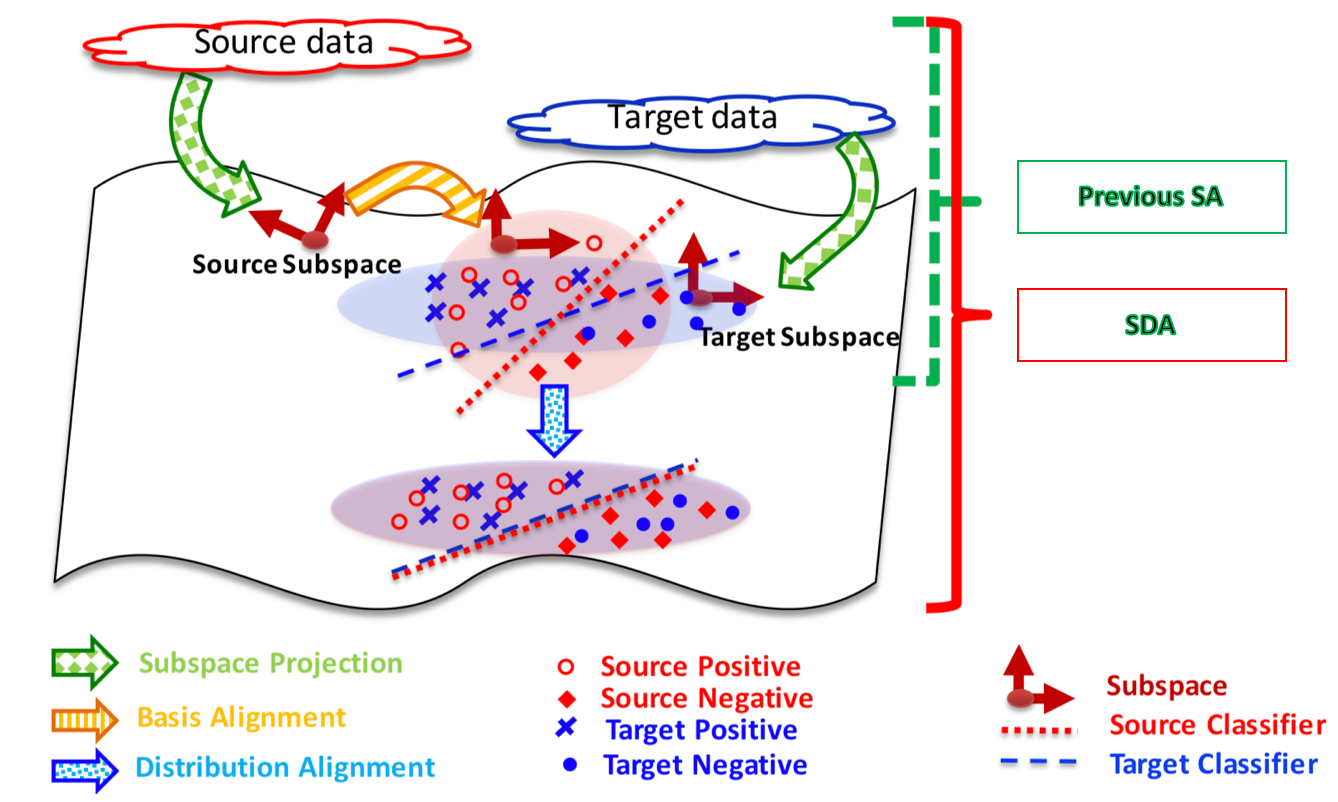}
\caption{The scheme of SDA model. The model considered the subspace alignment and distribution adaptation (image modified from \protect\cite{wang2018visual}). }
\label{fig:SDA}
\end{figure}

However, SA did not take the difference between the source distribution and the target distribution into account. Sun et al.~\cite{sun2015subspace} proposed the subspace distribution alignment, which can not only align the feature space but also align the distributions of domains. The SDA model improves the domain alignment via the distribution alignment. They first projected the labeled source-domain instances to the source subspace, then mapped to the target subspace, and finally mapped back to the target domain. 

One advantage of subspace-based methods is that the calculation is simple and efficient. Similarly, the linear correlation alignment (CORAL) minimized domain shift by aligning the second-order statistics of source and target distributions
\cite{sun2016return}; it solved the following optimization problem:

\begin{equation}\label{eq:CORAL}
\min_A ||C_{\hat{S}}-C_T||_F^2=\min_A ||A^TC_SA-C_T||_F^2, 
\end{equation}
where $A$ is the transformation matrix, $C_{\hat{S}}$ is the covariance of the transformed source features
$X_SA$. $C_S$, and $C_T$ are covariance matrices of source and the target domain, respectively. The main process of CORAL is updating the source data using its covariance followed by the ``re-coloring" of the target covariance matrix.

There are also approaches to minimize domain discrepancy based
on the spectral feature alignment using graph theory. Pan et al. proposed a spectral feature alignment (SFA)~\cite{pan2010cross} method. It can identify the domain-specific and domain-independent features in different domains and then aligns these domain-specific features by constructing a lower-dimensional feature representation.

\begin{figure}[t]
\centering
\includegraphics[scale=0.380]{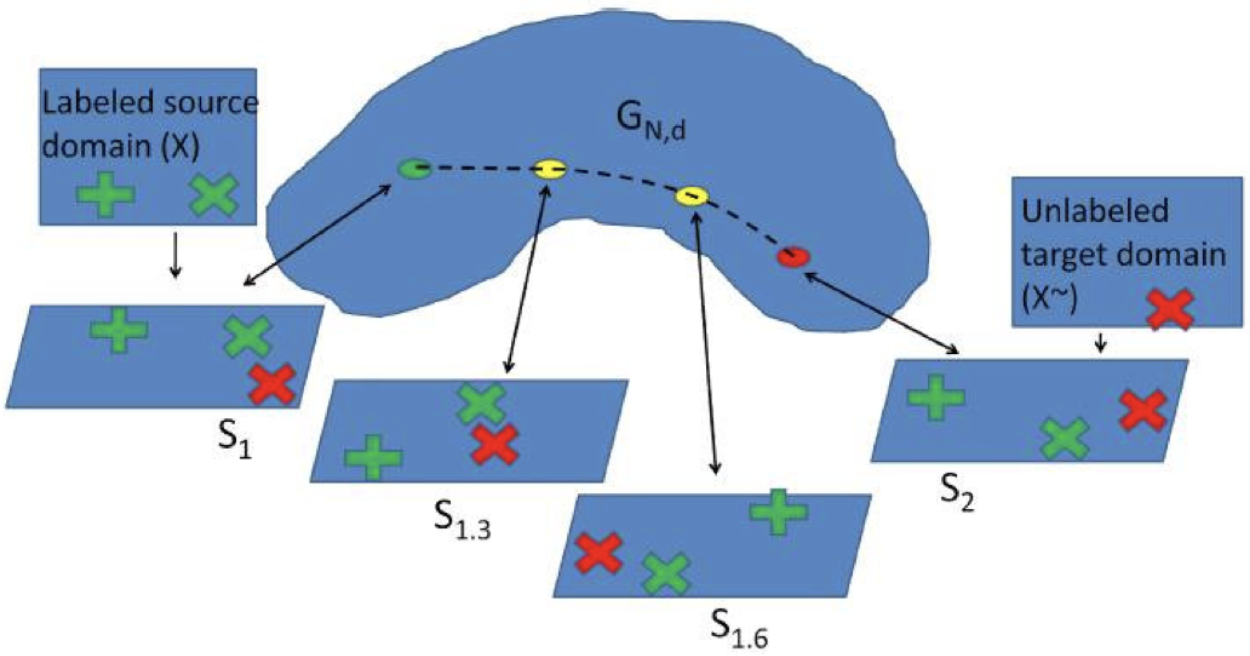}
\caption{The scheme of sampling geodesic flow (SGF) method (image from~\protect\cite{gopalan2011domain}). }
\label{fig:SGF}
\end{figure}

Manifold learning models aim to map the data on Riemannian manifold and reduce the distance of the two domains on the manifold. One of the earliest manifold learning methods is based on the Grassmannian manifold, which learns the intermediate features between the sub-source and the sub-target domain via a Grassmannian manifold.  Gopalan et al.\ \cite{gopalan2011domain} proposed a sampling geodesic flow (SGF) method to learn the intermediate features between the sub-source and the sub-target domain via the geodesic (shortest path) on Grassmannian manifold.
To obtain the samples between $\mathcal{X_S}$ and $\mathcal{X_T}$, sampling geodesic flow (SGF) consists of the following steps: 1) calculate the geodesic which starts from the source and ends with target domains on the Grassmannian in the subspace; 2) sample a given number of subspaces along the geodesic; 3) project the original feature vectors into samples' subspaces and utilize the results as new features; 4) reduce the dimensionality of the new features; and, 5) use the resulting new (reduced) feature vectors to train the classifiers and evaluate on target data.

However, SGF has several limitations. Gong et al.\ have noted that it is difficult to choose an optimal sampling strategy \cite{gong2012geodesic}.  Also, several basic parameters need to be adjusted: the sample size, the reduced dimension of the subspace, and how to represent original data using new samples.  Moreover, SGF has high time complexity, making sampling slow when many points are needed. 
\begin{figure}[h]
\centering
\includegraphics[scale=0.43]{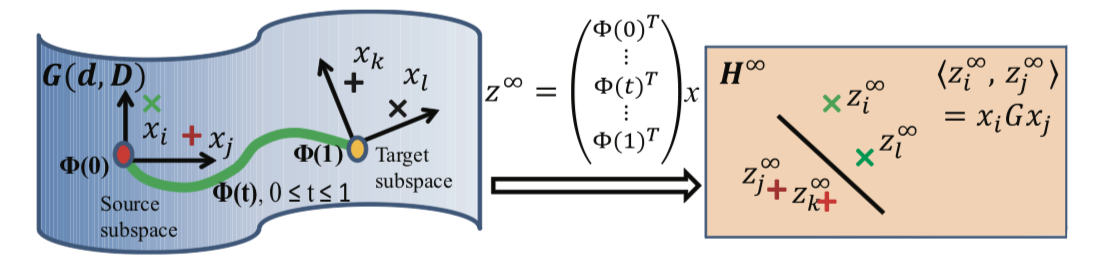}
\caption{The scheme of geodesic flow kernel (GFK) model. It considers all samples points on the geodesic (image from~\protect\cite{gong2012geodesic}). }
\label{fig:GFK}
\end{figure}

To overcome the limitations of unknown sampling size and subspace dimensionality, the geodesic flow kernel (GFK) was proposed by Gong et al.~\cite{gong2012geodesic}.  They integrated all samples along the ``geodesic" (the shortest distance between two points on the manifold), which is shown in the following equation.
\begin{equation}\label{eq:gfk}
    \int_{0}^{1} (\Phi(t)^T x_i)^T(\Phi(t)^T x_i)dt=x_i^TGx_i, 
\end{equation}
where $\Phi$ is the projection matrix.  The GFK model contained the following steps: 1) compute the optimal reduced dimensionality of the subspaces; 2) calculate the geodesic curve; 3) compute the geodesic flow kernel; and, 4) use the kernel to train a classifier with labeled data and test on unlabeled data.
However, dimensionality is a hyper-parameter of the GFK model; one needs to calculate the optimal dimensionality. In addition, it has the constraint that the size of dimensionality should be less than half of the minimum dimension of source and test data, which is $d < \frac{1}{2}\min (length(\mathcal{X}_\mathcal{S}^{'}), length(\mathcal{X}_\mathcal{T}^{'}))$, where $length$ refers to the number of features in the sub-source $\mathcal{X}_\mathcal{S}^{'}$ and sub-target $\mathcal{X}_\mathcal{T}^{'}$ domains. In addition, the GFK model will only work well if the dimensionality of each point is far larger than the number of total points.

\begin{figure}[t]
\centering
\includegraphics[scale=0.450]{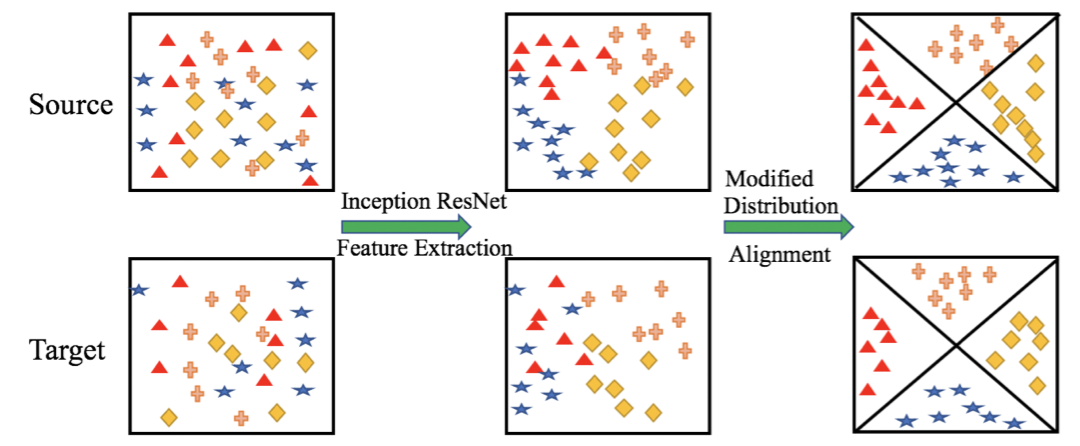}
\caption{The scheme of MDA model. Features are extracted from the last fully connected layer in InceptionResNetv2 model, and then align the distribution of learned features (image from \protect\cite{zhang2019modified}).}
\label{fig:MDA}
\end{figure}

However, none of these models explored the quality of the learned features, \textit{i.e.,} the geodesic path has not been verified. Zhang et al.,~\cite{zhang2019modified} found that the SGF method did not provide a correct way to sample the points along the geodesic. We also demonstrated that the ``geodesic" from the SGF model is not the true geodesic. They then extracted features from a pre-trained InceptionResNetv2 deep network. The deep features contained detailed information of the object, and the SGF-based manifold learning will destroy this information.  They also modified MEDA to form the modified distribution alignment (MDA) model, which improves the performance of the DA problem. The scheme of the MDA model is shown in Fig.~\ref{fig:MDA}. Later, They proposed a geodesic sampling on Riemannian manifolds (GSM)~\cite{zhang2019transductive} model to sample intermediate features along the correct geodesic. In the follow-up work, they proposed a subspace sampling demon (SSD)~\cite{zhang2020domain} approach to show the detailed shape deformations and utilize quantitative methods to evaluate learned features.  They also proposed a deep spherical manifold Gaussian kernel~\cite{zhang2021deep2} framework to map the source and target subspaces into a spherical manifold and reduce the discrepancy between them by embedding both extracted features and a Gaussian kernel.

\begin{figure}
\centering
\begin{subfigure}{0.3\textwidth}
\includegraphics[width=\linewidth]{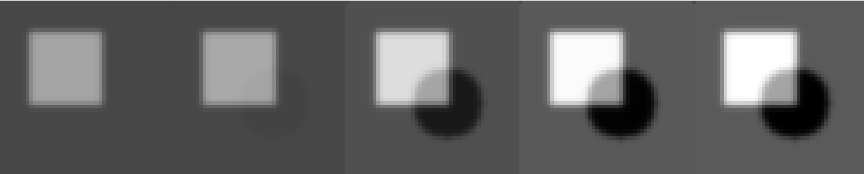}
\caption{SGF~\cite{gopalan2011domain}} \label{fig:ima}
\end{subfigure}
\begin{subfigure}{0.3\textwidth}
\includegraphics[width=\linewidth]{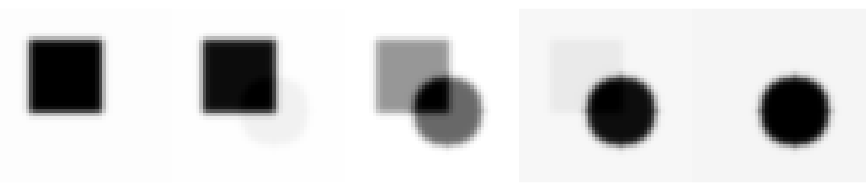}
\caption{GSM~\cite{zhang2019transductive}} \label{fig:imc}
\end{subfigure}
\begin{subfigure}{0.3\textwidth}
\includegraphics[width=\linewidth]{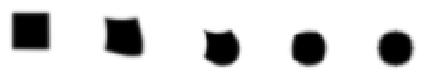}
\caption{ SSD~\cite{zhang2020domain} } \label{fig:imb}
\end{subfigure}
\caption{The comparison of sampling results between the two images (square and circle) with $t=0, \ 0.05, \ 0.5, \ 0.95, \ 1$. } \label{fig:im}
\end{figure}

\section{Deep Learning Methods}\label{sec:deep_me}
With the popularity of deep learning methods, deep neural networks have shown improved performance in transfer learning. Compared with traditional methods, deep transfer learning directly improves the learning effect on different tasks. Moreover, since deep learning directly learns from raw data, it has two advantages over traditional methods: automatically extracting more expressive features and meeting the end-to-end requirements in practical applications. 

Deep learning methods can be classified into homogeneous DA and heterogeneous DA. We focus on deep homogeneous DA. There are six categories of homogeneous deep UDA methods:  discrepancy-based, adversarial-based, pseudo-labeling-based, reconstruction-based, representation-based and attention-based methods as shown in Fig~\ref{fig:tox}. 

\begin{enumerate}
    \item Discrepancy-based: these methods minimize the distance between the source domain and the target domain using different statically defined distance functions.
    \item Adversarial-based: these methods identify the domain invariant features via two competing networks.
    \item Pseudo-labeling-based: these methods generate pseudo labels for the target domain to reduce the domain divergence.    
    \item Reconstruction-based: these methods map two domains into a shared domain while preserving domain specific features.  
    \item Representation-based: these methods utilize the trained network to extract intermediate representations as an input for a new network.
    \item Attention-based:  these methods pay attention to regions of interests (ROIs), which maintains shared information of both source domain and the target domain.
\end{enumerate}


\subsection{Discrepancy-based methods}
Discrepancy based methods are one of the most popular deep network models, and it aims to decrease the differences between the two domains and align data distributions. Different distance loss functions are usually added in the activation layers of networks.  Discrepancy based methods can be further divided into eight subgroups as shown in Fig.~\ref{fig:tox}. We review these different distance functions in the following subsections.

\subsubsection{Maximum Mean Discrepancy (MMD)}
Maximum Mean Discrepancy (MMD) is one of the most popular distances in minimizing a distance between two distributions, as shown in Eq.~(\ref{eq:mmd}). It measures the distributions as the squared distance between their embeddings in the reproducing kernel Hilbert space. MMD is also the equivalent to finding the RKHS mapping function, which
maximizes the difference in expectations between the two probability distributions in the following equation.
\begin{equation}
    MMD(\mathcal{D_S}, \mathcal{D_T}) = sup_{ f \in \mathcal{H}} ||  E_{\mathcal{X_S}}[f(\mathcal{X_S})] - E_{\mathcal{X_T}} [f(\mathcal{X_T} )]||_\mathcal{H}^2 ,
\end{equation}
where $E$ is the distribution expectation, and $f$ is a function or classifier in the deep neural networks.

Based on MMD, Tzeng et al.~\cite{tzeng2014deep} proposed a deep domain confusion (DDC) model; it minimized the following loss function:
\begin{equation}\label{eq:DDC}
\mathcal{L}=\mathcal{L}_C(\mathcal{X_S},\mathcal{Y_S})+\lambda MMD^2(\mathcal{X_S}, \mathcal{X_T}), 
\end{equation}
where $\mathcal{L}_C(\mathcal{X_S},\mathcal{Y_S})$ denotes the cross-entropy loss on the available labeled data ($\mathcal{X_S}$), and the ground truth labels ($\mathcal{Y_S}$), and $MMD^2(\mathcal{X_S}, \mathcal{X_T})$ denotes the distance between $\mathcal{X_S}$ and $\mathcal{X_T}$. The hyperparameter $\lambda$ determines robustness to confuse the domains. DDC model fixed the first seven layers and added the adaptation metric (MMD) in the eighth layer. Later, they extended the DDC model by introducing soft label distribution matching loss~\cite{tzeng2015simultaneous}.
\begin{figure}[h]
\centering
\includegraphics[scale=0.48]{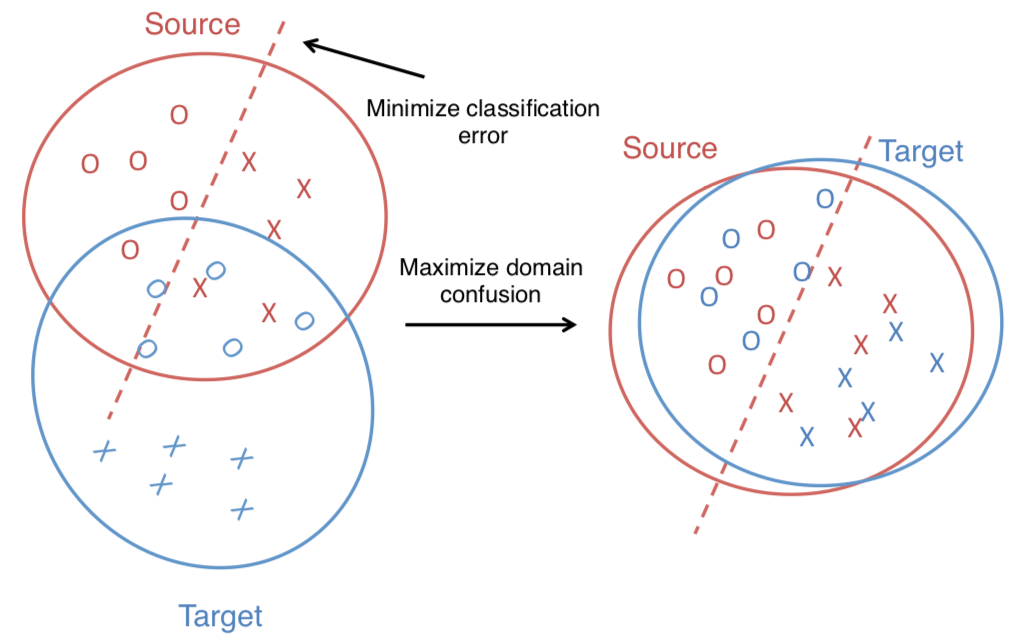}
\caption{The scheme of the deep domain confusion (DDC) model (image from~\protect\cite{tzeng2014deep}).}
\label{fig:DDC}
\end{figure}
Different from DDC, which used a single layer and linear MMD, the deep adaptation network (DAN)~\cite{long2015learning} model considered several MMDs between several layers and explored multiple kernels for adaptation of the deep representations. 
%

\begin{figure}[h]
\centering
\includegraphics[scale=0.450]{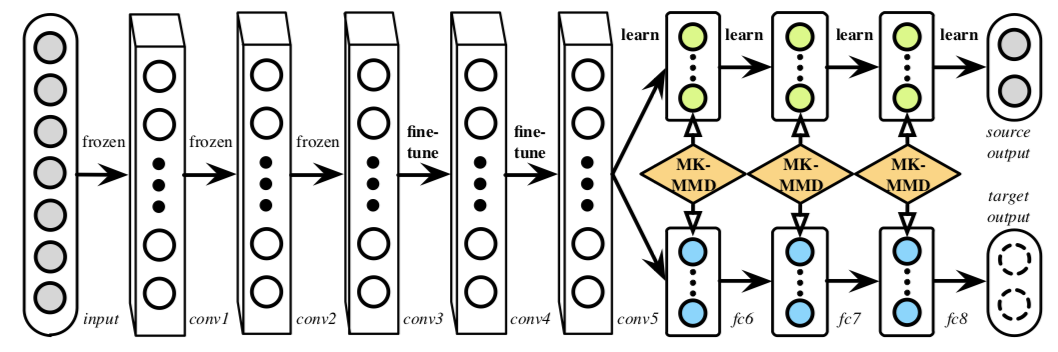}
\caption{The architecture of deep adaptation network (DAN) model.  The features are extracted from frozen (conv1–conv3) and fine-tuning (conv4–conv5) layers.  MK-MMD is adapted in fc6–fc8 layers (image from  \protect\cite{long2015learning}).}
\label{fig:DAN}
\end{figure}
Joint adaptation networks (JAN) \cite{long2017deep} further considered the joint distribution discrepancies (by using joint MMD (JMMD) criteria) of extracted features. In addition, residual transfer networks (RTN) \cite{long2016unsupervised} added a gated residual layer and relaxed the DAN classifier criteria. Yan et al.~\cite{yan2017mind} proposed a weighted MMD (WMMD) to construct the source distribution using the target domain to reduce the effect of class weight bias. Recently, the multi-representation adaptation network (MRAN)~\cite{zhu2019multi} extended  MMD to conditional MMD (CMMD) to reduce the differences between domains. Kang et al.~\cite{kang2019contrastive} extended MMD to the contrastive domain discrepancy loss. It can jointly optimize the intra-class distance and inter-class distance for improving the adaptation performance. 
Deng et al.~\cite{deng2020rethinking} considered triplet loss to align data distributions from domain-level and class-level. For aligning domain level, they utilized the JMMD metric to reduce the domain-level discrepancy, and similarity guided constraint (SGC) to reduce the class-level discrepancy.

\subsubsection{Correlation Alignment (CORAL)}
CORAL~\cite{sun2016deep} aims to align the second-order statistics (covariances)
between the cross-domain distributions. The Deep CORAL model extended the CORAL model into a deep architecture, and the loss function is defined in Eq.~(\ref{eq:JAN}).
\begin{equation}
\begin{aligned}\label{eq:JAN}
\mathcal{L}_{CORAL}=\frac{1}{4d^2}||C_\mathcal{S} - C_\mathcal{T}||_F^2,
\end{aligned}
\end{equation}
where $d$ is the feature dimensionality, $C_\mathcal{S}$ and $C_\mathcal{T}$ are the covariance matrices of the source data and the target data, and $||\cdot||_F^2$ denotes the squared matrix Frobenius norm. Mapped Correlation Alignment (MCA)~\cite{zhang2018unsupervised} projected covariances of different domains from Riemannian manifold to RKHS. It can learn a non-linear mapping via combining MCA loss and classification loss. Chen et al.~\cite{chen2019joint} introduced joint discriminative domain alignment (JDDA), which utilized CORAL loss, and applied a discriminative loss to form an instance-based and center-based discriminative learning scheme for DA. Rahman et al.\ proposed a model based on the alignment of second-order statistics (covariances) as well as maximized the mean discrepancy of the source and target data~\cite{rahman2020minimum}.
\begin{figure}[h]
\centering
\includegraphics[scale=0.22]{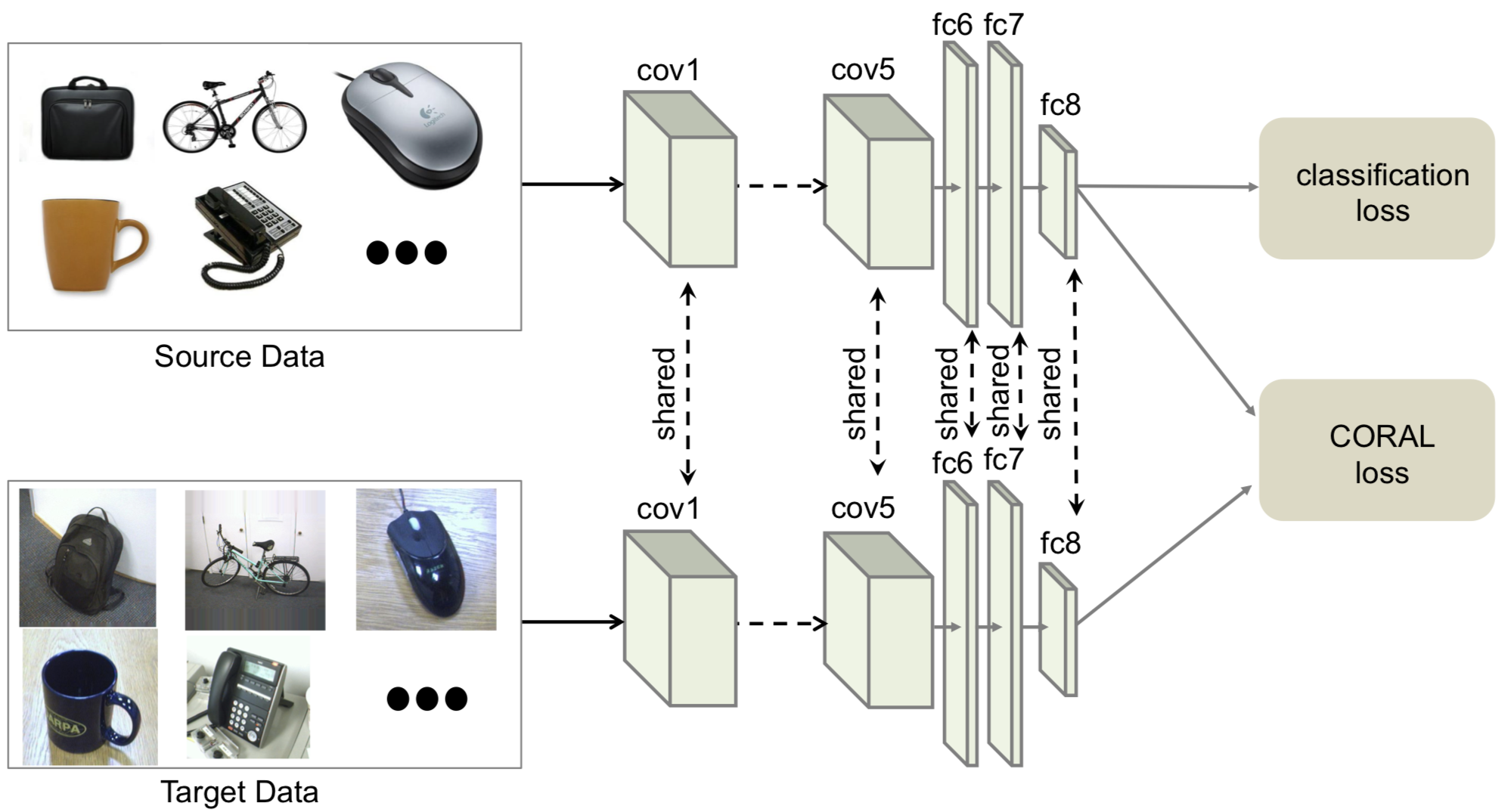}
\caption{The architecture of Deep CORAL model. It is based on a CNN with a classifier layer, which adds the CORAL loss on the fc8 layer of AlexNet (image from \protect\cite{sun2016deep}).}
\label{fig:Deep coral}
\end{figure}

\subsubsection{Kullback–Leibler Divergence (KL)}
Kullback–Leibler divergence (KL)~\cite{van2014renyi} aims to measure the distance between two distributions ($P(x)$ and $Q(x)$) as follows.
\begin{equation}
    D_{KL}(P||Q) = \sum_{x \in \mathcal{X}}P(x) \text{log} \frac{P(x)}{Q(x)}, 
\end{equation}
where $\mathcal{X}$ is the probability space, and KL divergence is an asymmetric distance: $D_{KL}(P||Q) \neq D_{KL}(Q||P)$. 
Zhuang et al.~\cite{zhuang2017supervised} proposed an approach
termed transfer learning with deep autoencoders (TLDA), which adopted two autoencoders for the source and the target domains via minimizing the KL divergence. Meng et al.~\cite{meng2018adversarial} also minimized the Kullback-Leibler divergence between the output distributions of the teacher and student models simultaneously to better align two domains.

\subsubsection{Jensen–Shannon Divergence}
Jensen–Shannon divergence (JSD)~\cite{fuglede2004jensen} is derived from KL divergence, and it is a symmetric distance. 
\begin{equation}
    JSD_{KL}(P||Q) = \frac{1}{2}  D_{KL}(P||M) + \frac{1}{2} D_{KL}(Q||M),
\end{equation}
where $M = \frac{1}{2}(P + Q)$.

Jiang et al.~\cite{jiang2020resource} proposed resource efficient domain adaptation (REDA) to distill transfer features across domain by minimizing the JSD between the probability predictions of the major classifier and the shallower classifiers.

\begin{figure}[h]
\centering
\includegraphics[scale=0.1]{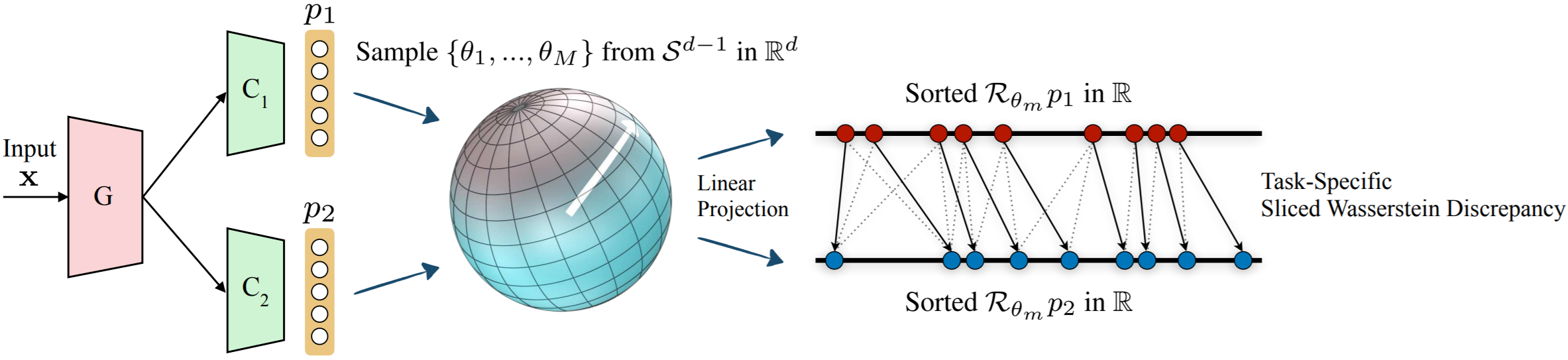}
\caption{The architecture of Sliced Wasserstein discrepancy (SWD) model (image from \protect\cite{lee2019sliced}).}
\label{fig:SWD}
\end{figure}

\subsubsection{Wasserstein Distance}
The Wasserstein metric~\cite{vallender1974calculation} is another discrepancy metric to measure the distance among the different domain samples. This distance is defined in a metric space ($M, \rho$), and $\rho(x_1,x_2)$ is the distance between two samples as shown in the following equation.
\begin{equation}
\begin{aligned}
    W(P(\mathcal{X_S}), P(\mathcal{X_T})) & = \{\inf_{\mu \in \Gamma(P(\mathcal{X_S}), P(\mathcal{X_T}))} \\ & \int \rho(x_1,x_2)^{p} d\mu(x_1,x_2)\}^{1/p}
\end{aligned}
\end{equation}
Damodaran et al.~\cite{bhushan2018deepjdot} jointly matched feature and label space distributions based on Wasserstein distance, and they not only learned
the new data representations aligned between the source and target domain,
but also simultaneously preserved the discriminative information.  Sliced Wasserstein discrepancy (SWD)~\cite{lee2019sliced} utilized the geometrical 1-Wasserstein distance as the discrepancy measure for obtaining the dissimilarity probability of source and target domains.

\subsubsection{Mutual Information}
Mutual Information (MI)~\cite{kraskov2004estimating} aims to find the similarity between two distributions in Eq.~(\ref{eq:ml}). 
\begin{equation}\label{eq:ml}
\begin{aligned}
    MI (P(\mathcal{X_S}), P(\mathcal{X_T}) ) =  \sum_{x_1 \in \mathcal{X_S}} & \sum_{x_2 \in \mathcal{X_T}} P(\mathcal{X_S}, \mathcal{X_T})  \\ & \text{log} \frac{P(\mathcal{X_S}, \mathcal{X_T})}{P(\mathcal{X_S}) P(\mathcal{X_T}) }, 
\end{aligned}    
\end{equation}

Gholami et al.~\cite{gholami2020unsupervised} employed a deep learning model to 
jointly maximize the mutual information between the domain labels and private (domain-specific) features while minimizing the mutual information between the domain labels and the shared (domain-invariant) features. Xie et al.~\cite{xie2020mi} disentangled the content features from domain information for both the source and translated images and then maximized the mutual information between the disentangled content features to preserve the
image-objects using a discriminator.

\subsubsection{Entropy Minimization}
Entropy minimization~\cite{grandvalet2005semi} aims to find the minimal entropy between distributions of two samples. Feature transfer network (FTN)~\cite{sohn2018unsupervised} first separated the transformed source domain and target domain using an
entropy minimization loss function to enhance the discriminative ability of FTNs in the target domain. Later, Roy et al.~\cite{roy2019unsupervised} proposed min-entropy consensus (MEC) method to jointly merge consistency loss and entropy loss to improve the domain
adaptation as shown in Eq.~\eqref{eq:em}.
\begin{equation}
\begin{aligned}\label{eq:em}
    & L^{t} (B_{1}^{t}, B_{2}^{t}) = \frac{1}{m} \sum_{i=1}^{m} l^{t} (x_{i}^{t_1}, x_{i}^{t_2} ), \\ \text{where} \ &
    l^{t} (x_{i}^{t_1}, x_{i}^{t_2} ) = -\frac{1}{2}\max_{y \in Y} (\log p (y|x_{i}^{t_1}) + \log p (y|x_{i}^{t_2}))
\end{aligned}
\end{equation}
where $x_{i}^{t_1} \in B_{1}^{t}$ and $x_{i}^{t_2} \in B_{2}^{t}$, and $B_{1}^{t}, B_{2}^{t}$ are two different target batches.

Mancini et al.~\cite{mancini2018boosting} further incorporated MEC loss with the multiple domain predictions on perturbations to achieve the consistency and reduce entropy for the perturbed domain predictions of the same input features.
\begin{figure}[h]
\centering
\includegraphics[scale=0.1]{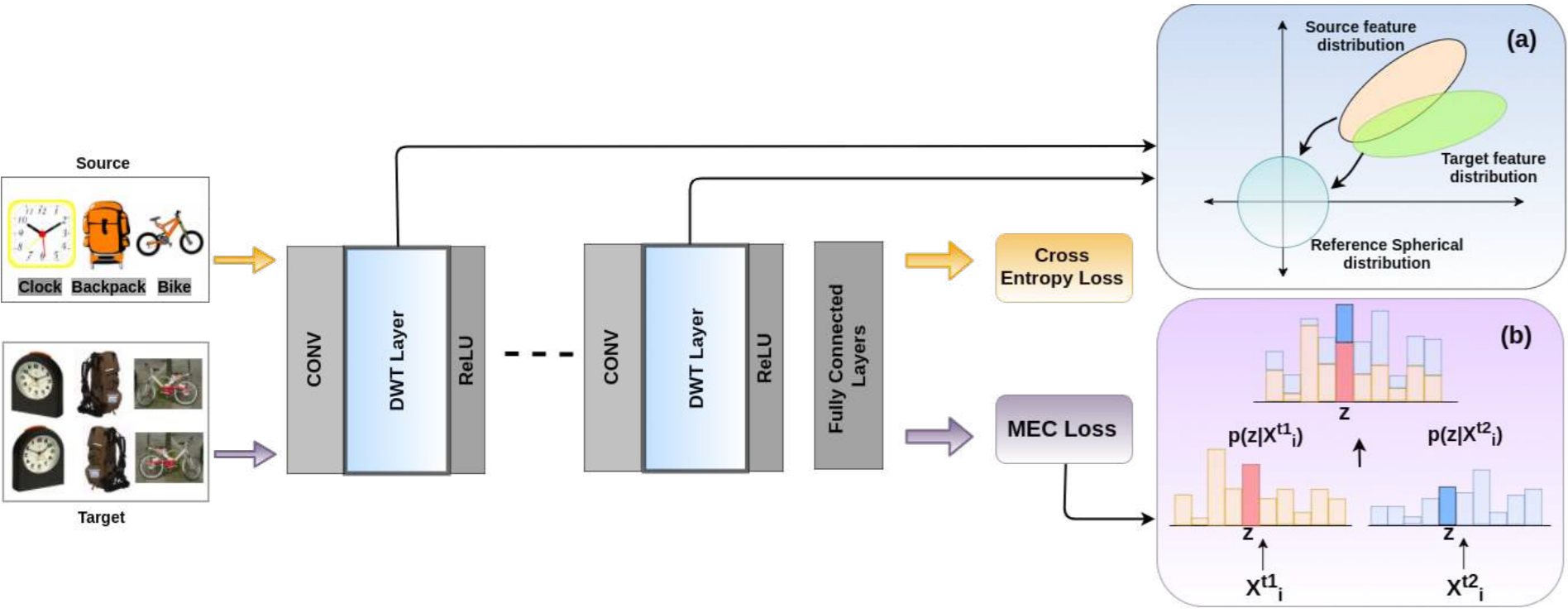}
\caption{The architecture of min-entropy consensus (MEC) (image from \protect\cite{roy2019unsupervised}).}
\label{fig:MEC}
\end{figure}

\subsubsection{Batch Normalization}
Batch Normalization (BN)~\cite{ioffe2015batch} has been widely used in deep networks to decrease the covariance shift. 

In multi-source DA~\cite{mancini2018boosting}, Mancini et al. extended batch normalization of DA layer to a new batch normalization layer (mDA-layer).  This mDA-layer can re-normalize the multi-modal feature distributions as shown in the following equation. 
\begin{equation}
mDA(x_i, w_i,\hat{\mu}, \hat{\sigma}) = \sum_{d \in D} w_{i,d} \frac{x_i - \hat{\mu}_d}{\sqrt{\hat{\sigma}_{d}^{2}+\epsilon}},
\end{equation}
where $w_i = [w_{i, d}]_{d \in D}$, $\hat{\mu} = [\hat{\mu}_d]_{d \in D}$ and $\hat{\sigma} = [\hat{\sigma}_{d}^{2}]_{d \in D}$. 

Li et al.~\cite{li2018adaptive} introduced an adaptive batch normalization (AdaBN) model to improve the generalization ability of the deep neural network. AdaBN can change the data of BN layers of the target domain via data of the source domain and also update the weights in CNN for DA. Change et al.~\cite{chang2019domain} proposed domain specific batch normalization (DSBN) based on multiple sets of BN layers. The DSBN can estimate the mean and variance of multiple domains, and it can capture the domain-specific features, and then the domain-invariant features can be better extracted from deep neural networks. 
\begin{figure}[h]
\centering
\includegraphics[scale=0.15]{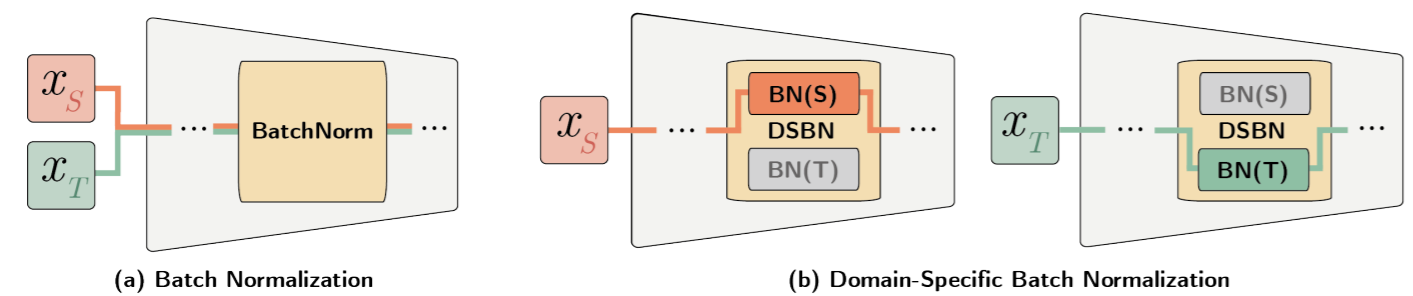}
\caption{The architecture of domain specific batch normalization (DSBN) (image from \protect\cite{chang2019domain}).}
\label{fig:DSBN}
\end{figure}

\subsubsection{Least Squares}
Least Squares~\cite{zhang2021deep33} aims to approach data distribution via estimating the slope and intercept in the latent space. Deep least squares alignment (DLSA)~\cite{zhang2021deep33} first propose to minimize the slop and intercept differences to realize domain divergence reduction with least squares estimation. They first minimized the marginal distribution as follows.
\begin{equation}\label{eq:mar}
  \mathcal{L_M} =  || \hat{a}_\mathcal{S} -\hat{a}_\mathcal{T}||_F^2 +  \gamma ||\hat{b}_\mathcal{S} - \hat{b}_\mathcal{T}||_F^2,
\end{equation}
where $\mathcal{M}$ denotes marginal distribution, $||\cdot||_F$ is the Frobenius norm, and $\gamma$ balances the scale between two terms. The first term enforces small differences of slope between two domains, while the second enforces small differences of intercept between two domains. They also minimized conditional distribution via reducing the categorical slop and intercept differences. 
\begin{figure}[h]
\centering
\includegraphics[scale=0.15]{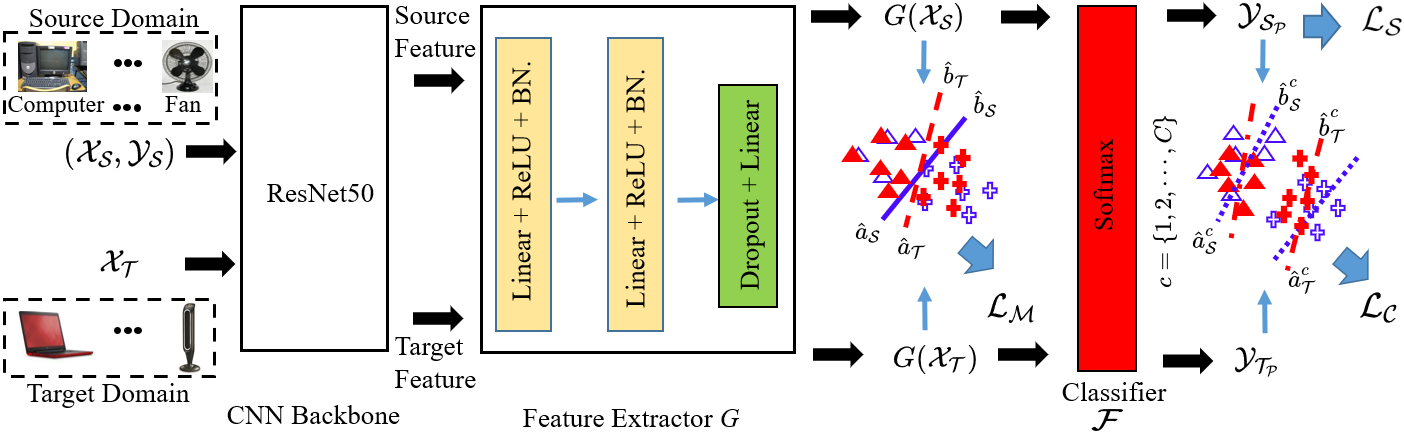}
\caption{The architecture of deep least squares alignment (DLSA) (image from \protect\cite{zhang2021deep33}).}
\label{fig:DLSA}
\end{figure}

\subsection{Adversarial-based methods}
Recently, adversarial-based methods have become an increasingly popular method to reduce domain discrepancy between different domains by using an adversarial objective. With the advent of generative adversarial networks (GAN)~\cite{goodfellow2014generative}, adversarial learning models have been found to be an impactful mechanism for identifying invariant representations in domain adaptation. Adversarial learning also contains a feature extractor and a domain discriminator. The domain discriminator aims to distinguish the source domain from the target domain, while the feature extractor aims to learn domain-invariant representations to fool the domain discriminator \cite{tzeng2017adversarial,long2018conditional,zhang2019domain,liu2019transferable,wei2021metaalign,ma2021adversarial,jing2021adversarial,akkaya2021self,zhang2021adversarial2,zhang2021adversarial_re}. The target domain error is expected to be minimized via minimax optimization.

\begin{figure}[h]
\centering
\includegraphics[scale=0.29]{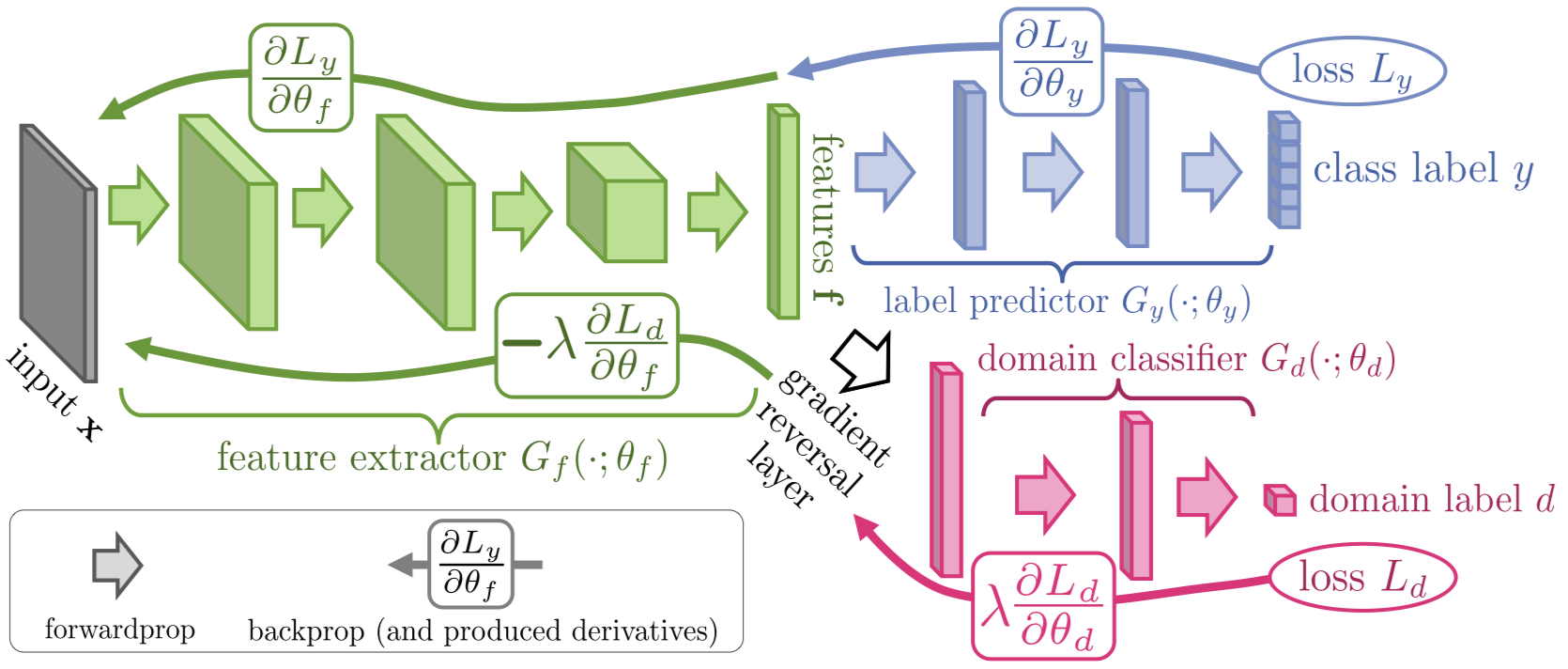}
\caption{The architecture of domain adversarial neural networks  (DANN) model. It includes a feature extractor (green), a label predictor (blue), and a domain classifier (pink)  (image from~\protect\cite{ganin2016domain}).}
\label{fig:DANN}
\end{figure}

The domain adversarial neural network (DANN) \cite{ganin2016domain} is one of the first adversarial methods for adversarial based DA. 
DANN considered a minimax loss to integrate a gradient reversal layer to promote the discrimination of source and target domains \cite{ganin2016domain}. Unsupervised DA is achieved by the gradient reversal layer that multiplies the gradient by a certain negative constant during the backpropagation-based training to ensure that the feature distributions over the two domains are made indistinguishable. The domain discriminator typically minimizes the binary cross-entropy loss as follows.
\begin{gather}\label{eq:all_caj1_caj}
\scalebox{0.99}{$
\begin{aligned}
    \mathcal{L_A}   =  &- \frac{1}{\mathcal{N_S}} \sum_{i=1}^{\mathcal{N_S}} \text{log} (1-D(\mathcal{X}_\mathcal{S}^i)) - \frac{1}{\mathcal{N_T}} \sum_{j=1}^{\mathcal{N_T}} \text{log} (D(\mathcal{X}_\mathcal{T}^j)) 
\end{aligned}$}
\end{gather}
\begin{figure}[h]
\centering
\includegraphics[scale=0.28]{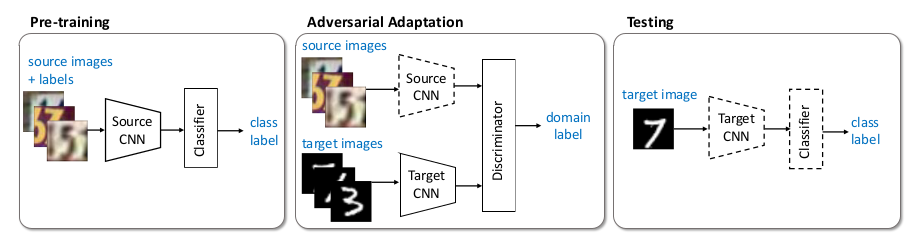}
\caption{The architecture of adversarial discriminative domain adaptation (ADDA) model. The dash lines represents fixed network parameters (image from~\protect\cite{tzeng2017adversarial}). }
\label{fig:ADDA}
\end{figure}

The adversarial discriminative domain adaptation (ADDA) uses an inverted label GAN loss to split the source and target domains, and features can be learned separately \cite{tzeng2017adversarial}. The coupled generative adversarial networks \cite{liu2016coupled} consisted of a series of GANs, and each of them can represent one of the domains. Cao et al. \cite{cao2018partial} proposed a partial transfer learning model. They noted that in the era of big data, we usually have a lot of source domain data. These source domain data are usually richer in categories than target domain data. For example, the image classifier based on ImageNet training must categorize thousands of categories. When we use it in practice, the target domain is often only a part of the categories. This leads to a problem: categories that exist only in the source domain will have a negative impact on label migration results.  The collaborative adversarial network (CAN) \cite{zhang2018collaborative} added several domain classifiers on multiple CNN feature extraction blocks on each domain classifier for DA. Chen et al.~\cite{chen2019joint} proposed joint domain alignment and discriminative feature learning. It benefits both domain alignment and final classification. Two discriminative feature learning methods are proposed (instance-based and center-based), which can guarantee the domain invariant features.

The joint adaptation network (JAN)~\cite{long2017deep} combined MMD with adversarial learning to align the joint distribution of multiple domain-specific layers across domains. Enhanced transport distance (ETD) measured domain discrepancy by establishing the transport distance of attention perception as the predictive feedback of iterative learning classifiers~\cite{li2020enhanced}. Cycle-consistent adversarial domain adaptation (CyCADA) proposed cycle-consistency loss to enforce local and global structural consistency between two domains~\cite{hoffman2018cycada}. To improve results, many methods utilize image-level adaptation (to maintain the consistency of images during training) to help feature alignment. Progressive domain adaptation~\cite{hsu2020progressive} combined feature alignment with image-level adaptation. They first adopted a model between source and intermediate domain via image translation. The transformed images have the same label mapped from the source domain and are treated as simulated training images for the target domain. Then, the intermediate and target domains are aligned. Zhang et al.~\cite{zhang2018collaborative}   reweighted the target samples, which can confuse the domain discriminator. The domain-symmetric network (SymNet) is a symmetrically designed source and target classifier based on an additional classifier. The proposed category-level loss can improve the domain-level loss by learning the invariant features between two domains~\cite{zhang2019domain}. 

Miyato et al.~\cite{miyato2018virtual} incorporated virtual adversarial training (VAT) in semi-supervised contexts to smooth the output distributions as a regularization of deep networks. Later, virtual adversarial domain adaptation (VADA) improved adversarial feature adaptation using VAT. It generated adversarial examples against only the source classifier and adapted on the target domain~\cite{shu2018dirt}. Unlike VADA method, transferable adversarial training (TAT) adversarially generated transferable examples that fit the gap between source and target domain~\cite{liu2019transferable}.  Xu et al.~\cite{xu2019adversarial} constructed a GAN-based architecture named adversarial domain adaptation with domain mixup (DM-ADA). It maps the two domains to a common potential distribution, and effectively transfers domain knowledge. Zhang et al.~\cite{zhang2020hybrid} introduced a hybrid adversarial network (HAN) to minimize the source classifier loss, conditional adversarial loss, and correlation alignment loss. A new adaptation layer was used to further improve the performance in the HAN model.

\subsection{Pseudo-labeling-based methods}
Pseudo-labeling is another technique to address UDA and also achieves substantial performance on multiple tasks. Pseudo-labeling typically generates pseudo labels for the target domain based on the predicted class probability~\cite{saito2017asymmetric,xie2018learning,zhang2018collaborative,chen2019progressive,zhang2020adversarial3}. Under such a regime, some target domain label information can be considered during training. In deep networks, the source classifier is usually treated as an initial pseudo labeler to generate the pseudo labels (and use them as if they were real labels). Different algorithms are proposed to obtain additional pseudo labels and promote distribution alignment between the two domains. 
\begin{figure}[h]
\centering
\includegraphics[scale=0.15]{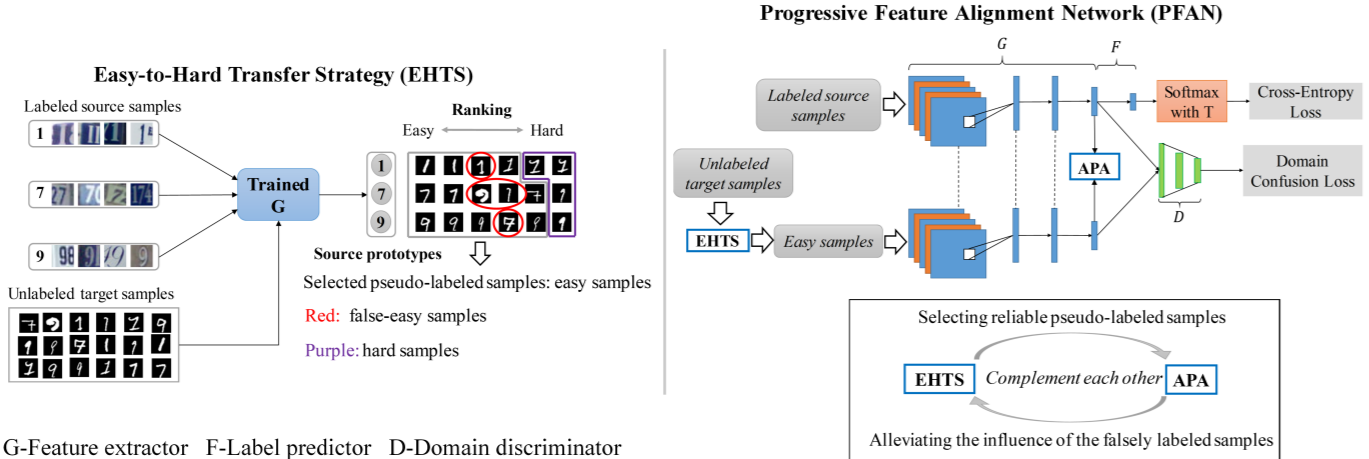}
\caption{The architecture of progressive feature alignment network (PFAN) (image from \protect\cite{chen2019progressive}).}
\label{fig:PFAN}
\end{figure}
An asymmetric tri-training method for UDA has been proposed by Saito et al.~to generate pseudo labels for target samples using two networks, and the third  can learn from them to obtain target discriminative representations~\cite{saito2017asymmetric}. Xie et al.~\cite{xie2018learning} proposed a  moving semantic transfer network (MSTN) to develop semantic matching and domain adversary losses to obtain pseudo labels. Zhang et al.~\cite{zhang2018collaborative} designed a new criterion to select pseudo-labeled target samples and developed an iterative approach called incremental collaborative and adversarial
network (iCAN), in which they select samples iteratively and retrain the network using the expanded training set. Progressive feature alignment network (PFAN)~\cite{chen2019progressive}  aligns the discriminative features across domains progressively and employs an easy-to-hard transfer strategy for iterative learning. Chang et al.~\cite{chang2019domain} proposed to combine the external UDA algorithm and the proposed domain-specific batch normalization to estimate the pseudo labels of samples in the target domain and more effectively learn the domain-specific features. Constrictive adaptation network (CAN) also employed batch normalization layers to capture the domain-specific distributions~\cite{kang2019contrastive}. Zhang et al.~\cite{zhang2020label} offers a label propagation with augmented anchors (A2LP) method to improve label propagation via generation of unlabeled virtual samples with high confidence label prediction.  Adversarial continuous learning in UDA (ACDA)~\cite{zhang2020adversarial3} increased
robustness by incorporating high-confidence samples from the target domain to the source domain. They further proposed a pre-trained features selection and recurrent pseudo-labeling (PRPL)~\cite{zhang2021efficient} model to continuously generate high-quality pseudo labels.

\begin{figure}[h]
\centering
\includegraphics[scale=0.25]{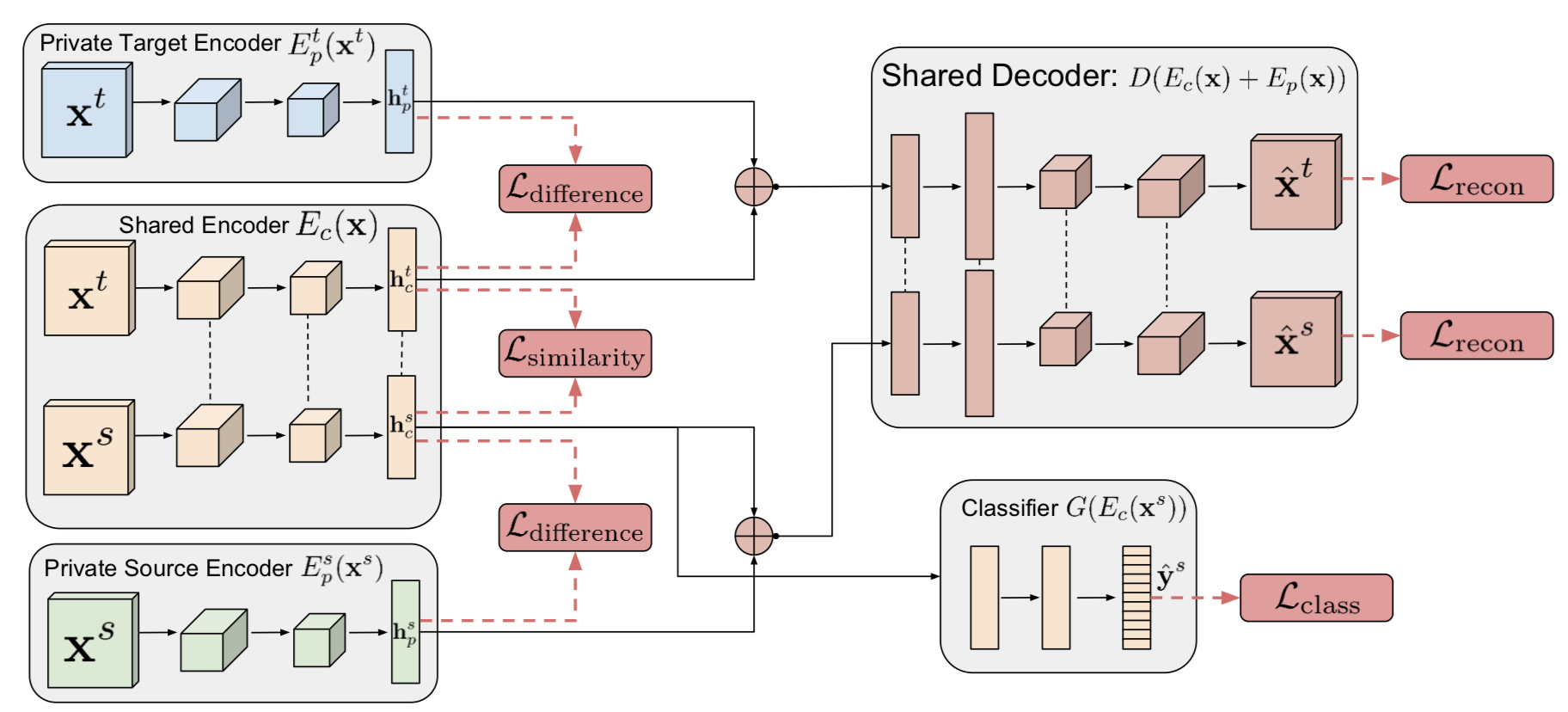}
\caption{The architecture of domain separation networks (DSN) model. It consists of four loss functions: $l_{class}$,  $l_{recon}$, $l_{difference}$ and $ l _{similarity}$  (image from~\protect\cite{bousmalis2016domain}). }
\label{fig:DSN}
\end{figure}

\subsection{Reconstruction-based methods}
Reconstruction based methods aim to reconstruct all domain samples to make better representations of domains, while preserving domain specific features.

Encoder-decoder style is one representative reconstruction based method. It first encodes input images to some hidden features by the encoder, then decodes these features back for reconstructed images by the decoder. The domain-invariant features are learned by a shared encoder while domain-specific features are preserved by reconstruction loss~\cite{ghifary2016deep}.  Stacked denoising autoencoders (SDA)~\cite{vincent2008extracting} is one of the first deep models for domain adaptation and aimed to find the common features between source and target domains via denoising autoencoders. 
The objective function is defined in Eq.~\ref{eq:auto}.
\begin{equation}\label{eq:auto}
\begin{aligned}
\theta^{\star}\theta^{'\star}    & =\arg \min _{\theta^{\star}\theta^{'\star}} \frac{1}{n}  \sum_{i=1}^{n} L (x^{(i)}, z^{(i)}) 
 \\ &=\arg \min _{\theta^{\star}\theta^{'\star}} \frac{1}{n}  \sum_{i=1}^{n} L (x^{(i)}, g_{\theta'}(f_{\theta} (x^{(i)}))),
\end{aligned}
\end{equation}
where $x$ is the input vector, $L$ is the loss function, which is squared error: $L (x, z) = ||x-z||^2$, $\theta$ is the parameter in the autoencoders,  and $f$ and $g$ are mapping functions.

To reduce the computational costs of SDA model, Chen et al.~\cite{chen2012marginalized} introduced a marginalized SDA (mSDA) model to denoise the marginal noise with a closed-form solution without using a stochastic gradient descent strategy. Multi-task autoencoder (MTAE)~\cite{ghifary2015domain} learned intra- and inter-domain reconstruction to represent domain invariances. Ghifary et al.~\cite{ghifary2016deep} proposed a deep reconstruction classification network (DRCN) to learn a shared encoding representation, which aims to minimize domain discrepancy. Zhang et al.~\cite{zhuang2015supervised} proposed transfer learning with deep auto-encoders using Kullback–Leibler divergence to reduce the discrepancy between the source and target distributions. Domain separation networks (DSN)~\cite{bousmalis2016domain} introduced the notion of a private subspace for each domain, which captures domain-specific properties, such as background and low-level image statistics. The shared subspace is enforced through the use of autoencoders and explicit loss functions, which can capture common features between the two domains. The loss function is defined as following:

\begin{equation}\label{eq:DSN}
\begin{aligned}
l=l_task+ \alpha l_{recon} +\beta l_{difference} + \gamma l _{similarity},
\end{aligned}
\end{equation}
where $l_{task}$ is the loss of the training, $l_{recon}$ is the loss of the reconstruction, $l_{difference}$ is the difference between public and private space, and $ l _{similarity}$ is the similarity of public space of source and target domain. The architecture of DSN is shown in Fig.\ref{fig:DSN}.

\subsection{Representation-based methods }
Representation based methods utilize the trained network to extract intermediate representations as an input for a new network. 

\begin{figure}[h]
\centering
\includegraphics[scale=0.4]{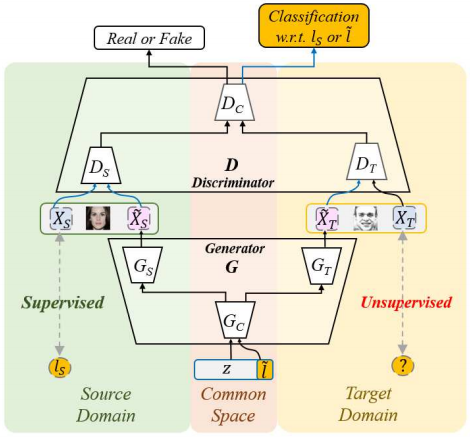}
\caption{The architecture of cross
domain representation disentangler (CDRD) model (image from~\protect\cite{liu2018detach}).}
\label{fig:CDRD}
\end{figure}

One common method is called disentanglement representation, which is based on class labels to gain invariant feature representation.  Cross domain representation disentangler (CDRD)~\cite{liu2018detach} bridged the labeled source domain and unlabeled target domain by jointly learning cross-domain feature representation disentanglement and adaptation.  The common space is optimized with both the labeled source domain and the unlabeled target domain. Hence, the shared weighted common space can bridge the gap between high and coarse-level representations of cross-domain data. Gonzalez et al.~\cite{gonzalez2018image} proposed an image-to-image translation for representation disentangling based on GANs and cross-domain autoencoders. They separated the internal representation into three parts: a shared part, which contains the domain-invariant features for two domains; and two exclusive parts, which contain the domain-specific features. Their model can be applied to multiple tasks, such as diverse sample generation, cross-domain retrieval, domain-specific image transfer, and interpolation. Liu et al.~\cite{liu2018unified} introduced a unified feature disentanglement network (UFDN) to learn domain-invariant representation from multiple domains for image translation and manipulation. Peng et al.~\cite{peng2019domain} minimized mutual information between domain-specific and domain-invariant features to pursue implicit domain-invariant features, which can improve the performance of the target domain. Gholami et al.~\cite{gholami2020unsupervised} presented a multi-target domain adaptation information-theoretic approach (MTDA-ITA) to find a shared latent space of all domains by simultaneously identifying the remaining private domain-specific factors. They utilized a unified information-theoretic approach to disentangle the shared and private information while establishing a connection between latent representations and the observed data. Their model can adapt from a single source to multiple target domains.  However, these disentanglement-based methods are still difficult to guarantee the full separation between domain-specific features and domain invariant features. Also, the reconstruction of these two features is less considered. Zhang et al.~\cite{zhang2021enhanced} propose an enhanced separable disentanglement (ESD) model. It can teach a disentangler to distill domain-specific and domain-invariant features from the two domains. They then applied feature separation maximization processes to enhance the disentangler to remove contamination between these two features. A reconstructor is used to recover original feature prototypes to further improve the performance of the model.
\begin{figure}[h]
\centering
\includegraphics[scale=0.15]{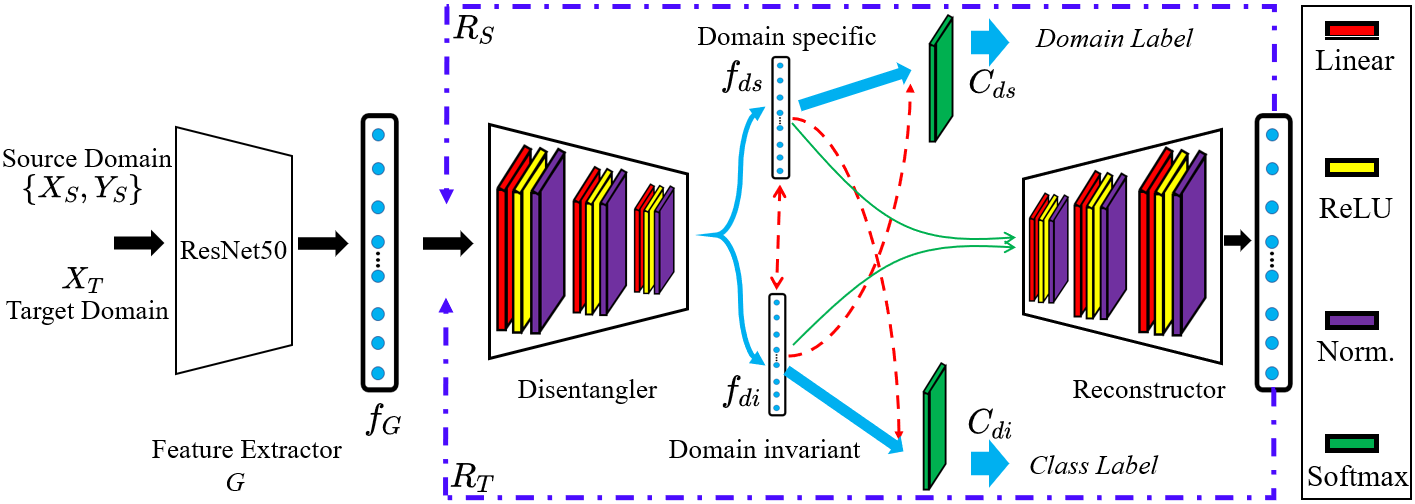}
\caption{The architecture of enhanced separable disentanglement (ESD) model (image from~\protect\cite{zhang2021enhanced}).}
\label{fig:ESD}
\end{figure}

\subsection{Attention-based methods }
Attention based methods pay attention to region of interests (ROIs) from the source domain to the target domain, which can make the deep neural networks focus on some spatial parts of both domains.

Wang et al.~\cite{wang2017residual} proposed a residual attention network (RAN), which added an attention mechanism in a convolutional neural network. RAN can generate attention-aware features via stacking attention modules. The attention module contains three key parameters: the number of pre-processing Residual Units before splitting into the trunk branch and mask branch, the number of Residual units in the trunk branch, and the number of Residual units between the adjacent pooling layer in the mask branch. However, RAN has the issue of negative local attention in transferring tasks. Later, the transferable attention for domain adaptation (TADA) model reduced the effects of negative transfer. It applied transferable global attention based on local attention. There are two types of complementary transferable attention: local attention can generate transferable regions by multiple region-level domain discriminators, and global attention can generate transferable images by image-level domain discriminator. 
\begin{figure}[t]
\centering
\includegraphics[scale=0.25]{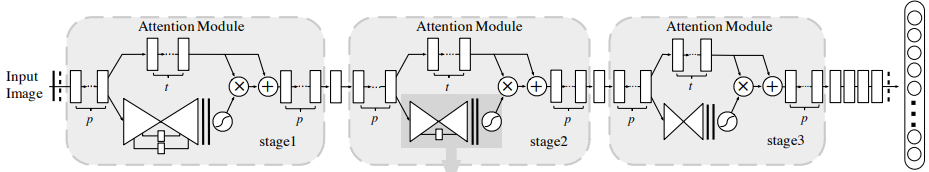}
\caption{The architecture of residual attention network (RAN) (image from~\protect\cite{wang2017residual}).}
\label{fig:RAN}
\end{figure}

Zhuo et al.~\cite{zhuo2017deep} presented a deep unsupervised convolutional domain adaptation (DUCDA) model, which consists of source classification loss and correlation alignment (CORAL) loss. The CORAL loss measured the discrepancy between attention maps for both source and target domains, and it was used on both convolutional layers and fully connected layers. Moon et al.~\cite{moon2017completely} proposed completely heterogeneous transfer
learning (CHTL) to filter and suppress irrelevant source samples using an attention mechanism and designed an unsupervised transfer loss to learn the knowledge between two domains. Kang et al.~\cite{kang2018deep} presented a deep adversarial attention alignment model, which transfers knowledge in all the convolutional layers via attention alignment. In addition, they estimated the posterior label distribution of the unlabeled domain, and they utilized category distribution to calculate the cross-entropy loss for training in improving predicting accuracy.

\section{Datasets \& SOTA results}\label{sec:results}

In this section, we list benchmark datasets for visual
domain adaptation. These datasets are important since they are widely used to evaluate the performance of domain adaptation models. Table~\ref{tab:data_des} summarized the statistics of eight benchmark datasets.

\subsection{Office + Caltech-10}\label{sec:OC10}
\begin{figure}[h]
\centering
\includegraphics[scale=0.29]{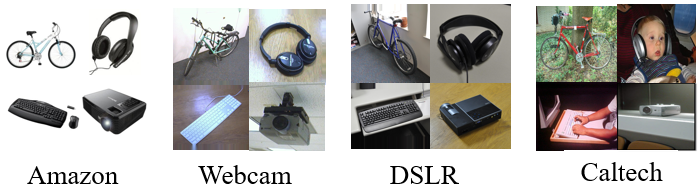}
\caption{Sample images from four categories across the four domains of the Office + Caltech-10 dataset (image from~\protect\cite{zhang2021unsupervised}).}
\label{fig:OC_im}
\end{figure}

This dataset~\cite{gong2012geodesic} is a standard benchmark for domain adaptation, which consists of Office 10 and Caltech 10  datasets. It contains 2,533 images in four domains in ten categories: Amazon, Webcam, DSLR, and Caltech. Amazon images are mostly from online merchants;  DSLR  and Webcam images are mostly from offices. Caltech images are from more real-world backgrounds.  Fig.~\ref{fig:OC_im} shows sample images from the Office + Caltech-10 dataset.

\subsection{Office-31}\label{sec:O31}
\begin{figure}[h]
\centering
\includegraphics[scale=0.29]{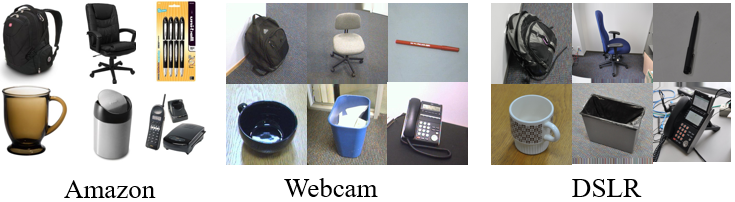}
\caption{Sample images from three domains of the Office-31 dataset. We select images from six categories (image from~\protect\cite{zhang2021unsupervised}).}
\label{fig:O31_im}
\end{figure}

Office-31 \cite{saenko2010adapting} is another benchmark dataset for domain adaptation, and it consists of 4,110 images in 31 classes from three domains: Amazon, which contains images from amazon.com; Webcam and DSLR, both contain images that are taken by a web camera or a digital SLR camera with different settings. Fig.~\ref{fig:O31_im} shows sample images from the Office-31 dataset.

\subsection{Office-Home}\label{sec:OH}
\begin{figure}[h]
\centering
\includegraphics[scale=0.29]{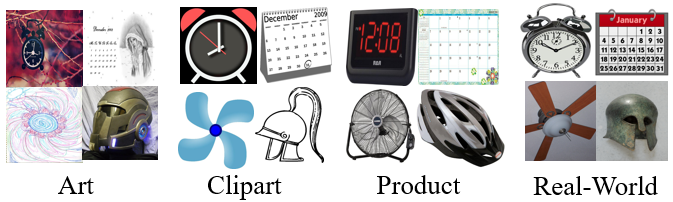}
\caption{Sample images from four domains of the Office-Home dataset. We only show images from four categories (image from~\protect\cite{zhang2021unsupervised}).}
\label{fig:OH_im}
\end{figure}

Office-Home \cite{venkateswara2017deep} contains 15,588 images in 65 categories across four domains. Specifically,  Art denotes artistic depictions for object images; Clipart describes picture collections of clipart; Product shows object images with a clear background and is similar to Amazon category in Office-31,  and Real-World represents object images collected with a regular camera. It is a challenging dataset since the domain divergence between different domains is larger.
Fig.~\ref{fig:OH_im} shows sample images from the Office-Home dataset.

\subsection{MNIST-USPS}

The MNIST-USPS dataset contains handwritten digit images and consists of the MNIST dataset~\cite{lecun1998gradient} and the US Postal (USPS) dataset~\cite{friedman2001elements}. Each dataset has ten categories. The MNIST dataset is derived from the National Institute of Standards and Technology (NIST) dataset. The MNIST dataset has 60,000 training samples and 10,000 test samples. The USPS dataset obtains recognized handwritten digits. The training set and the test set have 7291 and 2007 samples, respectively. 
\begin{figure}[h]
\centering
\includegraphics[scale=0.29]{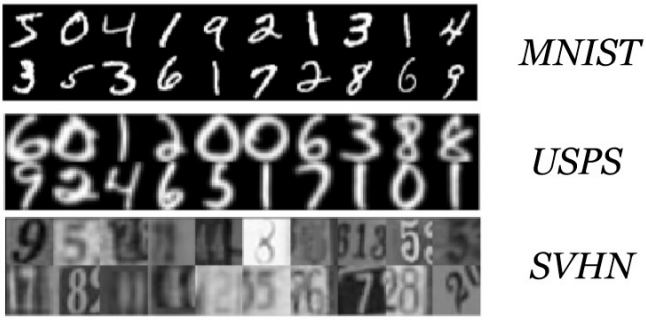}
\caption{Sample images from MNIST, USPS, and SVHN dataset (image from~\protect\cite{zhang2021unsupervised}).}
\label{fig:O31_UMV}
\end{figure}

\subsection{SVHN}

The SVHN dataset~\cite{netzer2011reading} has images from the street view house number from Google. This dataset is challenging due to changes in shape and textures, and extraneous numbers with the labeled image.  It has over 600,000 digit images with ten classes. Fig.~\ref{fig:O31_UMV} shows sample images from MNIST, USPS, and SVHN dataset, respectively.

\subsection{VisDA-2017}\label{sec:Vis}

\begin{figure}[h]
\centering
\includegraphics[scale=0.29]{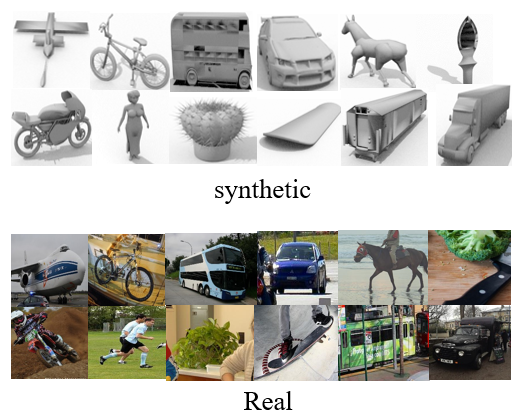}
\caption{Sample images of twelve classes from VisDA-2017 dataset (image from~\protect\cite{zhang2021unsupervised}).}
\label{fig:VisDA-2017}
\end{figure}

This dataset~\cite{peng2017visda} is closer to practical application scenarios and is a challenging dataset due to the significant domain-shift between the synthetic images (152,397 images from VisDA) and the real images (55,388 images from COCO) from 12 classes. The 12 classes are plane, bicycle, bus, car, horse, knife, motorcycle, person, plant, skateboard, train and truck as shown in Fig.~\ref{fig:VisDA-2017}.

\subsection{ImageCLEF-DA}

\begin{figure}[h]
\centering
\includegraphics[scale=0.29]{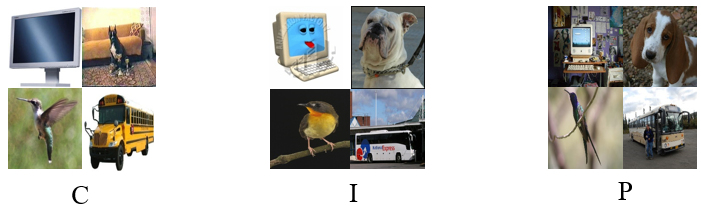}
\caption{Sample images from four domains of the ImageCLEF-DA dataset.  We only show images from four categories (image from~\protect\cite{zhang2021unsupervised}).}
\label{fig:ImageCLEF-DA}
\end{figure}

ImageCLEF-DA~\cite{ImageCLEF-DA} dataset is from ImageCLEF
2014 domain adaptation challenge. It contains three domains with a total of 600 images, which are formed by selecting images from three public datasets, including Caltech-256 (C), ImageNet ILSVRC
2012 (I), and Pascal VOC 2012 (P). Each domain
consists of 12 categories, and each category contains 50 images. 
The 12 classes are aeroplane, bike, bird,
boat, bottle, bus, car, dog, horse, monitor, motorbike, and people as shown in Fig.~\ref{fig:ImageCLEF-DA}.

\subsection{Amazon Reviews}

Amazon Reviews~\cite{blitzer2007biographies} is a multi-domain sentiment dataset that contains product reviews taken from Amazon.com of four domains (Books, Kitchen, Electronics and
DVDs). Each review in the four domains has a text and a
rating from zero to five.

\subsection{PIE}

The Carnegie Mellon University (CMU) Pose, Illumination, and Expression (PIE) database~\cite{Sim-2001-8175} contains 41,368 images of 68 people, where each person is represented under 13, 43, and 4, different poses, illuminations, and expressions, respectively. It has five subsets containing left pose,
up pose, down pose, front pose, and right pose.

\subsection{COIL20}
Columbia Object Image Library (COIL20)~\cite{Columbia} is a dataset of 1,440 normalized images with 20 object categories. The images are at pose intervals of 5 degrees.

\begin{table}[h!]
\small
\begin{center} 
\caption{Statistics of benchmark datasets}
\vspace{-0.3cm}
 \setlength{\tabcolsep}{+1.5mm}{
\begin{tabular}{cccccccc}
\hline \label{tab:data_des}
Dataset & $\#$ Sample  & $\#$ Class & Domain(s)  \\
\hline
Office-10  & 1,410  & 10 & A, W, D \\
Caltech-10 & 1,123   & 10 & C  \\
Office-31 & 4,110  &31 & A, W, D   \\
Office-Home & 15,588  &65 & Ar, Cl, Pr, Rw \\
MNIST-USPS-SVHN & 672,298 & 10 & M, U, S \\
VisDA-2017 & 207,785 & 12 & Synthetic, Real\\
ImageCLEF-DA & 1,800 & 12 & C, I, P  \\
Amazon Reviews & 8,000&2 & B, K, E, D  \\
\hline
\end{tabular}}
\end{center}
\end{table}

\begin{table}[h]
\footnotesize
\begin{center}
\caption{Statistics on PlantCLEF 2020 dataset}
\setlength{\tabcolsep}{+0.2mm}{
\begin{tabular}{rccccccccccccc}
\hline \label{tab:data}
Domain &  $\#$ Samples & $\#$ Classes \\
\hline
Herbarium (H) &  320,750 & 997 \\
Herbarium\_photo\_associations (A)& 1,816 &244\\
Photo (P)& 4,482  & 375 \\
Test (T) & 3,186   &- \\
 \hline
\end{tabular}}
\end{center}
\end{table}

\begin{figure}[h]
\centering
\includegraphics[scale=0.29]{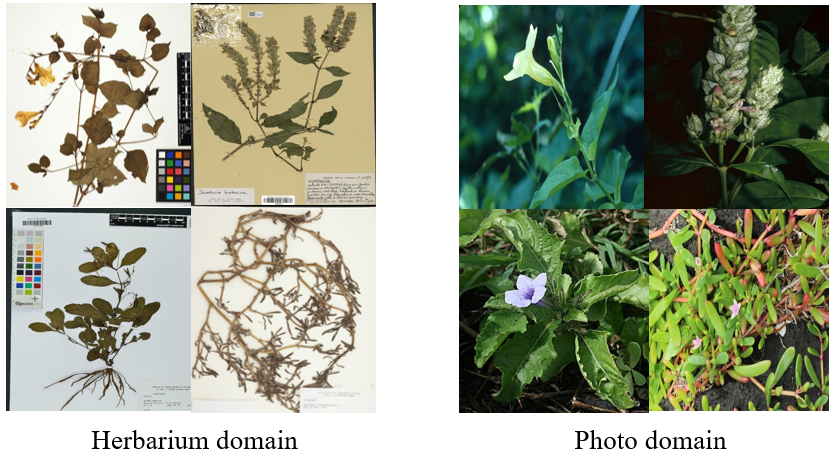}
\caption{Example images of the herbarium domain and photo domain. The large discrepancy between the two domains causes difficulty in improving the performance of the model (image from~\protect\cite{zhang2021unsupervised}). }
\label{fig:plant_eax}
\vspace{-0.3cm}
\end{figure}

\begin{table*}[h]
\small
\begin{center}
\captionsetup{font=small}
\caption{Accuracy (\%) on Office + Caltech-10 (based on ResNet50)}
\vspace{-0.3cm}
\setlength{\tabcolsep}{+0.9mm}{
\begin{tabular}{rccccccccccccc}
\hline \label{tab:OC+10}
Task & C$\shortrightarrow$A &  C$\shortrightarrow$W & C$\shortrightarrow$D & A$\shortrightarrow$C & A$\shortrightarrow$W & A$\shortrightarrow$D & W$\shortrightarrow$C & W$\shortrightarrow$A & W$\shortrightarrow$D & D$\shortrightarrow$C & D$\shortrightarrow$A & D$\shortrightarrow$W & \textbf{Ave.}\\
\hline
GSM~\cite{zhang2019transductive} & 96.0 &	95.9&	96.2&	94.6 &	89.5&	92.4&	94.1&	95.8&	\textbf{100}	& 93.9&	95.1&	98.6&	95.2\\
BDA~\cite{wang2017balanced} & 94.7&	93.2&	96.8&	89.0 &	87.8&	87.9&	86.5&	92.0	&99.4&	86.2&	92.3&	97.3&	91.9 \\
TJM~\cite{long2014transfer} & 94.7&	86.8&	86.6&	83.6&	82.7&	76.4&	88.2&	90.9&	98.7&	87.4&	92.5&	98.3&	88.9\\
JGSA~\cite{zhang2017joint} & 95.1&	97.6&	96.8&	93.9&	94.2&	96.2&	95.1&	95.9&	\textbf{100}	& 94.0 & 	\textbf{96.3} &	99.3&	96.2\\
MEDA~\cite{wang2018visual}	 & 96.3 &	98.3&	96.2&	94.6 &	99.0	&\textbf{100}	&94.8 &	\textbf{96.6}&	\textbf{100} & 93.6&	96.0	& 99.3&	97.0\\
DDC~\cite{tzeng2014deep}	&	91.9&	85.4&	88.8&		85.0&	86.1&	89.0&	78.0&	83.8&		\textbf{100}	&	79.0	&	87.1&	97.7	&	86.1\\
DCORAL~\cite{sun2016deep}	&	89.8&	97.3 &	91.0	&	91.9 &	\textbf{100}	& 90.5 &	83.7&	81.5&		90.1	& 88.6	&	80.1&	92.3&		89.7\\
DAN~\cite{long2015learning} 	&	92.0&	90.6&	89.3	&	84.1&	91.8&	91.7&	81.2&	92.1&		\textbf{100}	&	80.3	&	90.0 &	98.5&		90.1\\
RTN~\cite{long2016unsupervised}  &93.7 & 96.9 &94.2 &88.1 &95.2 & 95.5& 86.6& 92.5& \textbf{100} & 84.6& 93.8 & 99.2 &93.4 \\
MDDA~\cite{rahman2020minimum} &93.6 & 95.2 &93.4 &89.1 &95.7 & 96.6& 86.5&94.8 & \textbf{100} & 84.7& 94.7 & \textbf{99.4 }& 93.6\\
DLSA~\cite{zhang2021deep33}& \textbf{96.6	}& \textbf{98.6}	& \textbf{98.1}	& \textbf{95.4}	&98.9	&\textbf{100}	&\textbf{95.3} &	\textbf{96.6} &	\textbf{100}& \textbf{95.1} &	96.2	& 98.3  &	\textbf{97.4}\\
  \hline
\end{tabular}}
\end{center}
\end{table*}

\begin{table*}[h!]
\begin{center}
\small
\captionsetup{font=small}
\caption{Accuracy (\%) on Office-Home dataset (based on ResNet50)}
\vspace{-0.3cm}
\setlength{\tabcolsep}{+0.5mm}{
\begin{tabular}{rccccccccccccc}
\hline \label{tab:OH}
Task & Ar$\shortrightarrow$Cl &  Ar$\shortrightarrow$Pr & Ar$\shortrightarrow$Rw & Cl$\shortrightarrow$Ar & Cl$\shortrightarrow$Pr & Cl$\shortrightarrow$Rw & Pr$\shortrightarrow$Ar & Pr$\shortrightarrow$Cl & Pr$\shortrightarrow$Rw & Rw$\shortrightarrow$Ar & Rw$\shortrightarrow$Cl & Rw$\shortrightarrow$Pr & \textbf{Ave.}\\
\hline
GSM~\cite{zhang2019transductive} &49.4&	75.5&	80.2&	62.9&	70.6&	70.3&	65.6&	50.0&	80.8&	72.4&	50.4&	81.6&	67.5 \\
JGSA~\cite{zhang2017joint} &45.8&	73.7&	74.5&	52.3&	70.2&	71.4&	58.8&	47.3&	74.2&	60.4&	48.4&	76.8&	62.8\\
MEDA~\cite{wang2018visual} & 49.1&	75.6&	79.1&	66.7&	77.2&	75.8&	68.2&	50.4&	79.9&	71.9&	53.2&	82.0&	69.1\\
ResNet-50~\cite{he2016deep}&	34.9&	50.0&	58.0&	37.4&	41.9&	46.2&	38.5&	31.2&	60.4&	53.9&	41.2&	59.9&	46.1\\
DAN~\cite{long2015learning}	& 43.6	& 57.0& 	67.9& 	45.8& 	56.5& 	60.4& 	44.0& 	43.6& 	67.7& 	63.1& 	51.5& 	74.3& 	56.3\\
DANN~\cite{ghifary2014domain} 	& 45.6	& 59.3& 	70.1& 	47.0& 	58.5& 	60.9& 	46.1& 	43.7& 	68.5& 	63.2& 	51.8& 	76.8& 	57.6\\
JAN~\cite{long2017deep}		& 45.9& 	61.2& 	68.9& 	50.4& 	59.7& 	61.0& 	45.8& 	43.4& 	70.3& 	63.9& 	52.4& 	76.8& 	58.3\\
CDAN-M~\cite{long2018conditional}	& 50.6& 	65.9& 	73.4& 	55.7& 	62.7& 	64.2& 	51.8& 	49.1& 	74.5& 	68.2& 	56.9& 	80.7& 	62.8\\
TAT~\cite{liu2019transferable} &51.6  &69.5 & 75.4 & 59.4 & 69.5 & 68.6 & 59.5 &50.5 &76.8 &70.9 &56.6 &81.6 &65.8 \\
ETD~\cite{li2020enhanced} &51.3 & 71.9& \textbf{85.7}& 57.6 &69.2 &73.7 &57.8 &51.2 &79.3 &70.2 &57.5 &82.1 &67.3 \\
TADA~\cite{wang2019transferable} & 53.1 &72.3& 77.2& 59.1 &71.2& 72.1& 59.7& 53.1& 78.4 &72.4& \textbf{60.0} &82.9& 67.6 \\
SymNets~\cite{zhang2019domain} & 47.7 & 72.9 & 78.5 & 64.2  & 71.3  &74.2  & 64.2  & 48.8 &  79.5&  74.5 &52.6 & 82.7& 67.6 \\
DCAN~\cite{li2020domain} &54.5 &75.7 &81.2 &67.4 &74.0 &76.3 &67.4 &52.7 &80.6 &74.1 &59.1 &83.5 &70.5\\ 
RSDA~\cite{gu2020spherical} & 53.2 & 77.7 & 81.3 &  66.4 &  74.0 &  76.5 &  67.9 &  53.0 &  82.0 &  \textbf{75.8} &  57.8 &  85.4 &  70.9 \\
SPL~\cite{wang2020unsupervised} &54.5 &77.8 & 81.9  & 65.1 & 78.0 & \textbf{81.1} & 66.0 & 53.1 & \textbf{82.8} & 69.9 & 55.3 & \textbf{86.0} & 71.0 \\
ESD~\cite{zhang2021enhanced} & 53.2 &	 75.9  &	82.0  &	\textbf{68.4} &	 \textbf{79.3}  & 79.4 &  69.2 &	 54.8 &	81.9 &	74.6 &	56.2  &	83.8 & 71.6 \\
DLSA~\cite{zhang2021deep33}& 56.3 &	\textbf{79.4}  &	82.5  &	67.4 &	78.4  & 78.6 & \textbf{69.4} &	54.5 &	82.1 &	75.3 &	56.4  &	83.7 & 	71.7 \\ 
SHOT~\cite{liang2020we} &\textbf{57.1} & 78.1 & 81.5 & 68.0 & 78.2 & 78.1 & 67.4 & \textbf{54.9} &82.2 &73.3 & 58.8 &  84.3 & \textbf{71.8 }\\
\hline
\end{tabular}}
\end{center}
\end{table*}

\begin{table*}[h!]
\begin{center}
\small
\captionsetup{font=small}
\caption{Accuracy (\%) on VisDA-2017 dataset (based on ResNet101)}
\vspace{-0.3cm}
\setlength{\tabcolsep}{+1.1mm}{
\begin{tabular}{rccccccccccccc}
\hline \label{tab:VisDA}
Task &  plane & bcycl &  bus &  car &  horse&  knife &  mcycl & person &  plant & sktbrd & train & truck & \textbf{Ave.}\\
\hline
Source-only~\cite{he2016deep} &  55.1 & 53.3 & 61.9 & 59.1 &  80.6 & 17.9 &  79.7 & 31.2 &  81.0& 26.5& 73.5& 8.5 & 52.4 \\
DANN~\cite{ghifary2014domain} 	&81.9 &77.7& 82.8 &44.3& 81.2& 29.5 &65.1 &28.6 & 51.9 & 54.6 & 82.8 & 7.8 & 57.4\\
DAN~\cite{long2015learning}	& 87.1& 63.0& 76.5& 42.0 &90.3& 42.9 &85.9 &53.1& 49.7 &36.3& 85.8 &20.7 &61.1\\
 JAN~\cite{long2017deep}	&  75.7&  18.7&  82.3 & 86.3&  70.2 & 56.9&  80.5&  53.8 & 92.5 & 32.2 &  84.5 &  54.5 & 65.7 \\
 MCD~\cite{saito2018maximum}	&  87.0 & 60.9&  83.7&  64.0&  88.9&  79.6&  84.7&  76.9&  88.6&  40.3&  83.0&  25.8&  71.9\\
 DMP~\cite{luo2020unsupervised}  & 92.1  & 75.0  & 78.9 & 75.5 & 91.2 & 81.9 & 89.0 & 77.2 & 93.3 & 77.4 & 84.8 & 35.1 & 79.3 \\
 DADA~\cite{tang2020discriminative} &  92.9 & 74.2 &  82.5 &  65.0 &  90.9 &  93.8 &  87.2 &  74.2 &  89.9 &  71.5 &  86.5  & 48.7 &   79.8 \\
 STAR~\cite{lu2020stochastic} &  95.0 &  84.0 &  84.6 &  73.0 &  91.6 & 91.8&  85.9 & 78.4 &  94.4 &  84.7 & 87.0 & 42.2 &  82.7 \\
SHOT~{\scriptsize \cite{liang2020we}} &94.3 & 88.5 & 80.1 & 57.3 & 93.1 & 94.9 &80.7 & 80.3 & 91.5 &89.1 &86.3 &58.2 &82.9\\
DSGK~\cite{zhang2021deep}&  95.7 & 86.3 & \textbf{85.8} & 77.3 & 92.3 & 94.9 & 88.5 & \textbf{82.9} & 94.9 & 86.5 & 88.1 & 46.8 & 85.0 \\
 CAN~\cite{kang2019contrastive} &  \textbf{97.9} &  87.2 &  82.5 &  74.3 &  97.8 &  96.2 &  90.8 &  80.7 &  96.6 & 96.3 & 87.5 &  59.9 &  87.2 \\
DLSA~\cite{zhang2021deep33}&    96.9 &  \textbf{89.2}  &  85.4 &  \textbf{77.9} &  \textbf{98.3} &  \textbf{96.9} &  \textbf{91.3} & 82.6 &  \textbf{96.9} &  \textbf{96.5} & \textbf{88.3} &  \textbf{60.8} &  \textbf{88.4} \\
\hline
\end{tabular}}
\end{center}
\end{table*}

\begin{table}[h!]
\begin{center}
\small
\captionsetup{font=small}
\caption{Accuracy (\%) on Office-31 (ResNet50)}
\vspace{-0.3cm}
\setlength{\tabcolsep}{+1.1mm}{
\begin{tabular}{rccccccccccccc}
\hline \label{tab:O31}
Task & A$\shortrightarrow$W &  A$\shortrightarrow$D & W$\shortrightarrow$A & W$\shortrightarrow$D & D$\shortrightarrow$A & D$\shortrightarrow$W  & \textbf{Ave.}\\
\hline
GSM~{\scriptsize \cite{zhang2019transductive}} &85.9&	84.1&	75.5&	97.2&	73.6&	95.6&	85.3 \\
BDA~\cite{wang2017balanced}	& 77.0   &79.3 &  70.3  & 97.0  & 68.0 &  93.2 &  80.8\\
JGSA~{\scriptsize \cite{zhang2017joint}} &89.1&	91.0&	77.9&	\textbf{100} &	77.6&	98.2&	89.0 \\
MEDA~{\scriptsize \cite{wang2018visual}} &91.7&	89.2&	77.2&	97.4&	76.5&	96.2&	88.0 \\
RTN~{\scriptsize \cite{long2016unsupervised}} &	84.5 &	77.5 &	64.8 &	99.4 &	66.2 &	96.8 &	81.6 \\
ADDA~{\scriptsize \cite{tzeng2017adversarial}}	&86.2	&77.8  &68.9	&98.4 &	69.5 &	96.2	& 82.9\\
JAN~{\scriptsize \cite{long2017deep}}&	85.4 &	84.7	&70.0 &	99.8	&68.6 &	97.4 &	84.3\\
DMRL~{\scriptsize \cite{wu2020dual}} &90.8 &93.4 & 71.2  &\textbf{100} &73.0  &99.0  & 87.9 \\
TAT~{\scriptsize \cite{liu2019transferable}} & 92.5 & 93.2 & 73.1 & \textbf{100}& 73.1 & \textbf{99.3} & 88.4 \\
TADA~{\scriptsize \cite{wang2019transferable}} &94.3 & 91.6  & 73.0  &99.8  & 72.9 & 98.7 & 88.4 \\
{\scriptsize SymNets~\cite{zhang2019domain}\par}& 90.8& 93.9  &72.5  & \textbf{100} & 74.6 & 98.8& 88.4 \\
SHOT~{\scriptsize \cite{liang2020we}} &90.1  & 94.0  & 74.3 & 99.9 & 74.7 & 98.4 & 88.6 \\
SPL~{\scriptsize \cite{wang2020unsupervised}} &  92.7 & 93.0  & 76.8 & 99.8 & 76.4 & 98.7  & 89.6 \\
CAN~{\scriptsize \cite{kang2019contrastive}} & 94.5 & 95.0  &77.0   &99.8  & 78.0 & 99.1 &  90.6 \\
RSDA~{\scriptsize \cite{gu2020spherical}} & \textbf{96.1}  & 95.8  & 78.9 & \textbf{100} & 77.4  & \textbf{99.3} & 91.3 \\
DLSA~\cite{zhang2021deep33} & 95.2 &	\textbf{96.2} &	\textbf{80.4} &	99.2&	\textbf{80.7} &	98.0 &	\textbf{91.6}	\\
\hline
\end{tabular}}
\end{center}
\end{table}

\subsection{PlantCLEF 2020}
This dataset is a large-scale dataset of the PlantCLEF 2020 task~\cite{plantclef2020}. Fig.~\ref{fig:plant_eax} shows some challenging images in this dataset. Tab.~\ref{tab:data} lists the statistics on PlantCLEF 2021 dataset. The herbarium domain contains 320,750 images in 997 species, and the number of images in different species are unbalanced. This dataset consists of herbarium sheets whereas the test set will be composed of field pictures. The validation set consists of two domains herbarium\_photo\_associations and photos. Herbarium\_photo\_associations domain includes 1,816 images from 244 species. This domain contains both herbarium sheets and field pictures for a subset of species, which enables learning a mapping between the herbarium sheets domain and the field pictures domain. Another photo domain has 4,482 images from 375 species and images are from plant pictures in the field, which is similar to the test dataset. The test dataset contains 3,186 unlabeled images. Due to the significant difference between herbarium and real photos, it is extremely difficult to identify the correct class~\cite{zhang2020adversarial1,zhang2021weighted}.

\subsection{State-of-the-art results of image recognition}
As shown in Tab.~\ref{tab:OC+10}-Tab.~\ref{tab:O31}, we provide the results of four benchmark datasets (Office + Caltech-10, Office-31, Office-Home and VisDA-2017). In this experiment, C $\shortrightarrow$ A  means learning from existing domain C, and transferring knowledge to classify domain A. These results indicate that deep learning-based methods usually achieve better performance than traditional methods. However, some traditional methods (\cite{wang2018visual,zhang2019modified}) observe higher accuracy than some deep learning-based methods. This is mainly because the extracted features are from pre-trained deep neural networks. Therefore, there is a trend of combining traditional based methods with deep learning features. Also, deep learning models with pseudo-labeling techniques achieve promising results.

\section{Conclusions}
In this survey, we first introduce some basic notation for unsupervised domain adaptation, then review existing research in the context of UDA and describe benchmark datasets with some state-of-the-art performance. We focus on two categories of image recognition methods: traditional methods and deep learning based methods. 

Traditional methods rely on different feature extraction techniques to better represent images of the two domains. We discuss these methods from three directions: feature selection, distribution alignment, and subspace learning. Specifically, we illustrate three settings of distribution alignment: marginal, conditional, and joint distribution alignment. 

We present the deep learning based UDA from six directions:  discrepancy-based, adversarial-based, pseudo-labeling-based, reconstruction-based, representation-based, and attention-based methods. Specifically, we review eight different discrepancy based methods: maximum mean discrepancy, correlation alignment, Kullback–Leibler divergence, Jensen–Shannon divergence,  Wasserstein distance, mutual information, entropy minimization, batch normalization, and least squares estimation.
    
Although both traditional and deep learning-based methods have been proposed to solve the domain shift issue, they both have some limitations. Traditional methods heavily rely on the extracted deep features from well-trained neural networks to achieve better performance. Deep learning-based methods usually take a long computation time to train the images from scratch. In real-world applications, how to better extract deep features from images and design incremental and online UDA algorithms can be promising directions.

\bibliographystyle{unsrt}
\bibliography{ijcai19}

\end{document}